\documentclass{article}

\usepackage{PRIMEarxiv}

\usepackage[utf8]{inputenc} % allow utf-8 input
\usepackage[T1]{fontenc}    % use 8-bit T1 fonts
\usepackage{hyperref}       % hyperlinks
\usepackage{url}            % simple URL typesetting
\usepackage{booktabs}       % professional-quality tables
\usepackage{amsfonts}       % blackboard math symbols
\usepackage{nicefrac}       % compact symbols for 1/2, etc.
\usepackage{microtype}      % microtypography
\usepackage{lipsum}
\usepackage{graphicx}
\usepackage{natbib}
\usepackage{amsmath}
\usepackage{wrapfig}
\usepackage{array}
\usepackage{listings}
\usepackage{xcolor}
\usepackage{caption}
\usepackage{authblk}

\lstdefinestyle{bwstyle}{
  basicstyle=\ttfamily\small,
  numbers=left,
  numbersep=8pt,
  numberstyle=\tiny\color{black},
  breaklines=true,
  breakatwhitespace=false,
  columns=fullflexible,
  showstringspaces=false,
  keepspaces=true,
  frame=lines,
  rulecolor=\color{black},
  tabsize=2,
  aboveskip=0.8\baselineskip,
  belowskip=0.8\baselineskip,
  keywordstyle=\bfseries,
  commentstyle=\itshape,
  stringstyle=\itshape,
}
\DeclareCaptionType{listing2}[Listing][List of Listings]
  
\title{LLM-driven design of physics-constrained constitutive models: two agents are better than one}

\author[1]{Marius Tacke}
\author[2]{Matthias Busch}
\author[2]{Kian Abdolazizi}
\author[1]{Jonas Eichinger}
\author[3]{Kevin Linka}
\author[2,4,5]{Roland Aydin}
\author[1,2]{Christian Cyron}

\affil[1]{Helmholtz-Zentrum Hereon, Geesthacht, Germany}
\affil[2]{Hamburg University of Technology, Hamburg, Germany}
\affil[3]{RWTH Aachen University, Aachen, Germany}
\affil[4]{Saarland University, Saarbrücken, Germany}
\affil[5]{German Center for Artificial Intelligence, Kaiserslautern, Germany}

\begin{document}
\maketitle

\begin{abstract}
Developing constitutive models that capture how materials deform under load traditionally requires years of specialized expertise in continuum mechanics, machine learning, and scientific programming. Large language models (LLMs) have recently been shown to lower this barrier by generating constitutive models on demand, but existing single-agent pipelines lack systematic checks that the resulting models respect fundamental physical laws. To close this gap, we introduce the first multi-agent LLM-driven approach for constitutive model generation: a Creator agent proposes a model tailored to the data, while an Inspector agent critically audits each proposal against nine physical constraints and returns it for refinement whenever a violation is detected. We demonstrate this concept with constitutive artificial neural networks (CANNs) and benchmark it on brain tissue and rubber as isotropic materials, and on porcine skin tissue as a transversely isotropic material with a preferred fiber direction, using two different LLM backbones (Claude Opus 4.7 and Kimi K2.5). Whether a generated model satisfies the physical constraints is assessed numerically, by probing each constraint across a broad sample of deformation states, rotations, and perturbation directions. Adding the Inspector raises the share of exported models that pass all these checks from 90\% to 95\% for Opus and from 47\% to 60\% for Kimi. In addition, the generated models are on par with or even surpass expert-designed models in accuracy, extrapolate reliably beyond the training data, and generalize remarkably well to unseen loading paths. Separating generation from inspection thus turns LLM-driven constitutive modeling into a substantially more trustworthy process. The paradigm is deliberately technique-agnostic and scales automatically with advances in LLM capability, opening a promising path toward automated, physics-aware model discovery.
\end{abstract}

\section{Introduction}\label{sec_Introduction}

\begin{figure}[h]
  \centering
  \includegraphics[width=\linewidth]{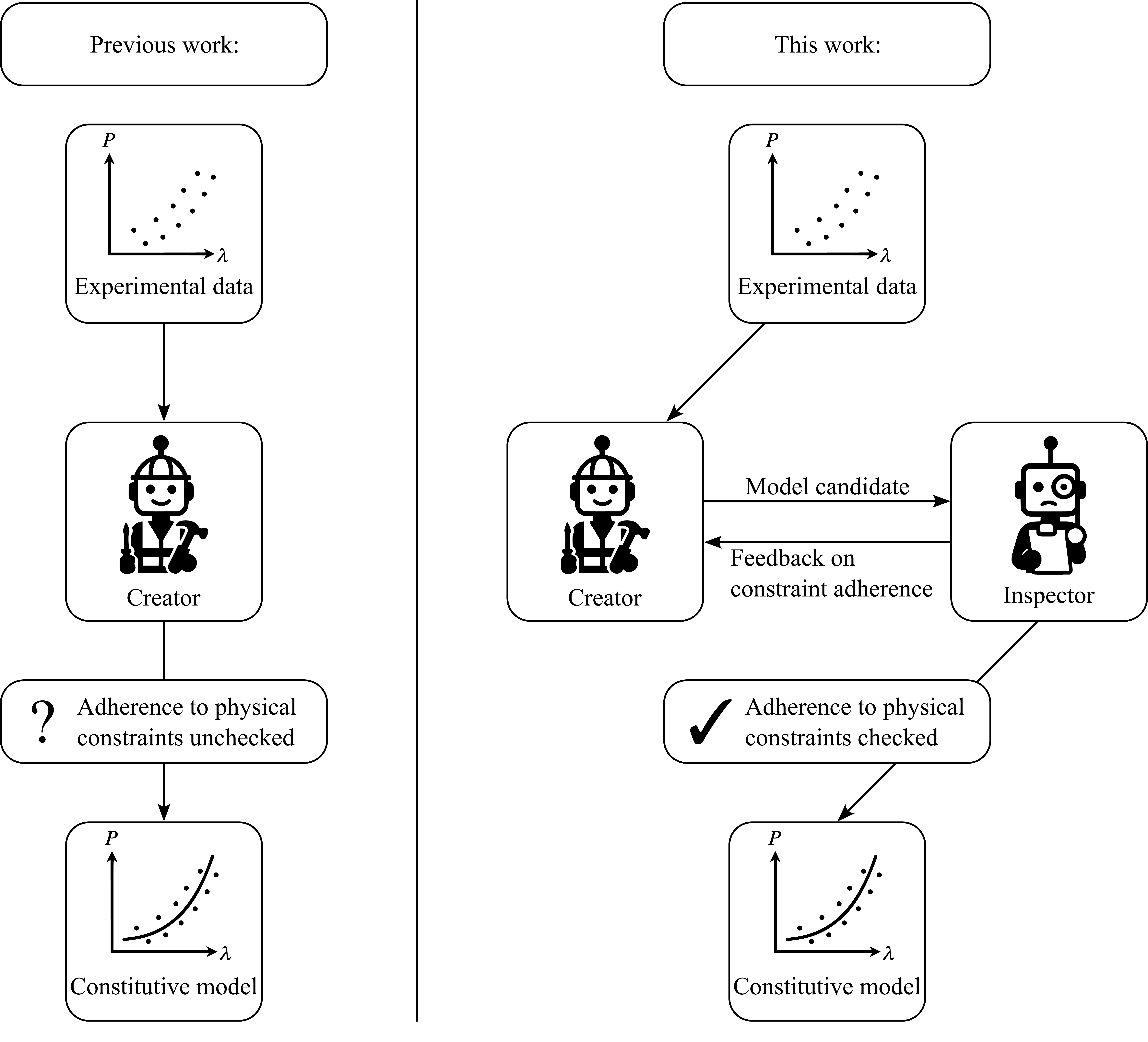}
  \caption{Previous LLM-based approaches to constitutive modeling rely on a single Creator agent that proposes a model and exports it directly (left). Instead, herein we propose a combination of two specialized agents (right): a Creator agent that proposes a constitutive model with a focus on physical admissibility, and an Inspector agent that critically audits the proposal against physical constraints. Only models that pass the Inspector are exported.}
  \label{fig_architecture_genncann_plus_inspector}
\end{figure}

Materials are typically characterized in the laboratory under relatively simple loading conditions, most often uniaxial and occasionally biaxial or shear. In service, however, they are subjected to far more complex loading. Constitutive models bridge this gap by providing a mathematical framework that is calibrated against simple experiments yet capable of predicting material response under more general loading conditions.

Traditionally, scientists have developed theory-driven constitutive approaches: the mathematical form of the model was chosen a priori from continuum mechanics, thermodynamics, and physical intuition, and then adjusted by hand to reproduce a limited set of experiments \cite{hooke1678potentia, maxwell1867iv, boltzmann1874theorie, prandtl1925spannungsverteilung, reuss1930berucksichtigung}. For hyperelasticity, this philosophy led to strain-energy formulations for rubber-like materials, in which investigators postulated scalar energy functions of strain invariants and differentiated them to obtain stresses \cite{mooney1940theory, rivlin1948largefundamental, rivlin1948largefurther, treloar1944stress, valanis1967strain, ogden1972large}. Such hand-crafted formulations are elegant and physically grounded but inherently limited. Developing and validating a new model typically requires years of specialized analytical effort, and the complexity of the resulting expressions is bounded by what a modeler can track by hand — leaving little room to capture the full richness of real material behavior.

With growing computational power and larger experimental and simulated datasets, data-driven approaches have emerged that bypass handcrafted functional forms altogether. Early efforts can broadly be characterized as black-box. Spline interpolations replace any explicit ansatz with piecewise polynomials and let the measurements speak directly through the interpolant \cite{sussman2009model, latorre2013extension, latorre2014you, xu2015nonlinear}. Distance-minimizing data-driven mechanics reformulates boundary value problems as a search for admissible mechanical states closest to a material database, circumventing the constitutive equation entirely \cite{kirchdoerfer2016data, carrara2020data}. Most influentially, artificial neural networks have served as universal approximators of stress–strain mappings, starting with the seminal work of Ghaboussi et al.\ in the early 1990s \cite{ghaboussi1991knowledge} and later integrated more tightly into finite element solvers \cite{hashash2004numerical}. While these methods offer remarkable flexibility, they share three well-documented shortcomings: they extrapolate poorly beyond the training data, they easily violate basic physical requirements such as objectivity, material symmetry, or thermodynamic consistency, and the resulting models are essentially opaque, providing little insight into the underlying deformation mechanisms \cite{linden2023neural, fuhg2024review, hao2022physics, abdolazizi2025constitutive}.

To overcome these limitations, a third class of hybrid or grey-box models has emerged that combines the flexibility of neural networks with physical priors of varying strength, differing both in their architecture and in the level of interpretability they provide \cite{fuhg2024review, hao2022physics}.  One line of work are constitutive artificial neural networks (CANNs), which embed strain invariants and admissible functional building blocks directly into the network topology so that the resulting strain energy satisfies physical constraints by construction, and which have been extended from hyperelasticity to viscoelasticity, inelasticity and growth–remodeling and applied to a wide range of soft materials including brain tissue, skin, meat and rubber-like elastomers \cite{linka2021constitutive, linka2023new, pierre2023principal, abdolazizi2024viscoelastic, holthusen2024theory, holthusen2025generalized, holthusen2025automated, linka2023automated, linka2023automatedmodel, pierre2023discovering, dal2023data, ehret2022variations, wiesheier2024versatile, peirlinck2025democratizing, mcculloch2024sparse}. A closely related line of work are physics-augmented neural networks (PANNs), which target the same set of constitutive conditions but employ multilayered input-convex architectures combined with analytical correction terms, offering additional expressive flexibility \cite{linden2023neural, klein2022polyconvex, as2022mechanics, kalina2025neural, friedrichs2026precise, geuken2026modeling, tac2022datadrivenmodeling, tac2022datadriventissue, tacc2023data}. A complementary family of grey-box approaches relies on sparse regression and symbolic discovery to identify compact analytical laws from libraries of physically motivated terms, ranging from genetic programming \cite{koza1993genetic, abdusalamov2023automatic} and sparse identification of nonlinear dynamics \cite{brunton2016discovering} to the EUCLID family that exploits full-field measurements and equilibrium constraints \cite{flaschel2021unsupervised, flaschel2022discovering, joshi2022bayesian, thakolkaran2022nn}. More recent work explores spline-based Kolmogorov–Arnold Networks (KANs) \cite{kolmogorov1956, liu2024kan, liu2024kan2}, which have been transferred to constitutive modeling as constitutive KANs (CKANs) that combine the predictive flexibility of neural networks with symbolic post-processing for enhanced interpretability \cite{abdolazizi2025constitutive, ji2026inelastic}. In a related spline-based approach, B-spline constitutive neural networks recover the strain energy of incompressible isotropic materials from tension–compression data \cite{lee2025constitutive}. Further hybrid concepts include model–data augmentation strategies \cite{tushar2025fusion} and physics-informed couplings to phase-field fracture \cite{dammass2025neural}. A complementary line of work incorporates physical information probabilistically rather than architecturally: physics-informed neural networks penalize violations of the governing equations in the loss \cite{raissi2019physics, karniadakis2021physics}, Bayesian variants thereof additionally quantify uncertainty \cite{yang2021bpinn, linka2022bayesian}, stochastic hyperelastic models construct maximum-entropy priors whose realizations satisfy admissibility requirements such as polyconvexity or the Baker--Ericksen inequalities \cite{staber2017stochastic, mihai2018stochastic}, and Bayesian model discovery with uncertainty quantification has been demonstrated for CANNs and EUCLID \cite{linka2024discovering, joshi2022bayesian}. These approaches enforce physical structure in a soft, probabilistic sense and quantify uncertainty alongside; here, in contrast, we require the physical constraints to hold deterministically.

In the present work, we build on the family of neural-network-based approaches, in particular CANNs and PANNs. A central concern shared by all of them is the rigorous enforcement of the mathematical and physical conditions that any admissible hyperelastic constitutive law must satisfy: thermodynamic consistency, stress tensor symmetry, objectivity, material symmetry, polyconvexity (and the implied ellipticity that guarantees material stability), the volumetric growth condition, normalization of energy and stress in the reference configuration, and non-negativity of the strain energy \cite{neff2015exponentiated, zee1983ordinary, upadhyay2019thermodynamics, holzapfel2002nonlinear}. A particularly systematic treatment is provided by Linden et al.\ \cite{linden2023neural}, who derive a network architecture that fulfills all of these conditions in an exact, by-construction manner for compressible isotropic and transversely isotropic hyperelasticity, building on architectural ingredients such as input-convex networks for polyconvexity \cite{klein2022polyconvex, as2022mechanics, geuken2026modeling}, analytical growth and normalization terms \cite{kalina2025neural}, stresses computed as gradients of a learned potential for thermodynamic consistency \cite{gonzalez2019thermodynamically, linka2023new, vlassis2021sobolev}, generalized structure tensors for anisotropy \cite{kalina2025neural}, and partially monotonic spline activations \cite{fritsch1980monotone, polo2025monokan} for KAN-based formulations \cite{abdolazizi2025constitutive}.

While the approaches reviewed above continue to automate ever larger parts of the constitutive modeling pipeline, their effective use still demands substantial expertise in continuum mechanics, machine learning, and scientific programming. Large language models (LLMs) have recently emerged as a complementary paradigm that further lowers this barrier by synthesizing domain knowledge, proposing hypotheses, and generating executable artifacts from natural-language descriptions \cite{nejjar2025llms, wysocki2024llm, liu2025beyond}. One line of work leverages LLMs for the discovery of symbolic scientific laws, where the model proposes equation skeletons that are then calibrated and iteratively refined through evolutionary search, retrieval augmentation, self-reflection, chain-of-thought prompting, or LLM-based plausibility scoring \cite{shojaee2025llm, wang2025drsr, zhang2025rag, sharlin2024context, taskin2026knowledge}, with analogous strategies transferred to the discovery of simulation-ready constitutive laws \cite{wu2026engineering}. A broader thrust employs LLMs as code generators that assemble full simulation or optimization pipelines from plain-language specifications, spanning surrogate modeling for engineering optimization \cite{rios2024large, hao2024large}, PDE solvers and physics-informed surrogates \cite{wuwu2025pinnsagent, li2025codepde}, graph algorithms and materials-property prediction \cite{verma2025grail, huang2025codegenerated}, chemical-process modeling \cite{heyer2025automated}, molecular dynamics \cite{shi2025fine}, CFD \cite{pandey2025openfoamgpt}, and, in solid mechanics, the generation of UMAT subroutines for rate-dependent plasticity \cite{gu2026large}. Beyond pure code generation, multimodal and tool-using systems couple LLMs with specialized numerical or experimental back ends for materials-property prediction, accelerated materials discovery, and autonomous experimentation \cite{tang2026multimodal, boyar2025llm, bazgir2025matagent, boiko2023autonomous, jia2024llmatdesign}.

A recurring pattern in this literature is a bilevel structure in which an LLM-driven outer loop proposes candidates, while an inner numerical loop calibrates and evaluates them \cite{chen2024llms, pandey2025openfoamgpt}. The scientific generative agent (SGA) generalizes this idea to open-ended hypothesis generation \cite{ma2024llm}, and the constitutive scientific generative agent (CSGA) specializes it to constitutive modeling by injecting assumptions such as isotropy, incompressibility, and an invariant basis; on stress–strain benchmarks it outperforms SGA but remains clearly less accurate than purpose-built CANNs \cite{tacke2025constitutive}. Most recently, Tacke et al. \cite{tacke2025automating} bridged this gap by letting an LLM generate task-specific CANNs (GenCANNs) on demand, reaching accuracy on par with, and in several cases exceeding, human-designed CANNs. Their framework, however, leaves open whether the generated networks actually satisfy the physical constraints that hand-crafted CANNs fulfill by construction \cite{linden2023neural}. This caveat is fundamental rather than incidental: by-construction adherence is a property of the model class but is inherited only by correctly implemented instances of it, and once the implementation is delegated to an LLM, its correctness can no longer be assumed but must be checked.

A parallel strand of the LLM literature distributes the task across specialized LLM instances that communicate, critique, and refine each other's outputs. Structured debate and self-consistency among multiple LLMs consistently improve factuality and reasoning over single-model chain-of-thought \cite{du2024improving, gridach2025agentic}, and rigid role assignments can be replaced by dynamic task decomposition and automatic agent instantiation \cite{wang2025tdag, chen2024autoagents, tang2025autoagent, liu2024a, nettem2025agentflow, yuan2025evoagent}. The recurring design principle, and the one we adopt here, is the contrast between generative agents, which propose code, hypotheses, or candidate solutions, and inspecting agents, which critique, verify, and refine these outputs \cite{gridach2025agentic, chen2024autoagents, liu2024a}.

The scientific-discovery community has embraced this paradigm, with agent teams now orchestrating large portions of the research loop and consistently outperforming strong single-LLM baselines on novelty, symbolic accuracy, and end-to-end autonomy \cite{lu2024ai, swanson2025virtual, boiko2023autonomous, su2025many, yang2025moosechem, xia2025sr, hu2026multi, yang2026think}. In materials and mechanics, applications range from bioinspired materials design and protein discovery to literature mining and robotic synthesis, and, closest to the present setting, collaborative finite-element code generation for continuum mechanics \cite{ghafarollahi2025sciagents, ghafarollahi2024protagents, chaudhari2026modular, lin2025reshaping, shi2026knowledge, pantiukhin2025accelerating, ni2024mechagents}. Across these studies, the multi-agent advantage is most pronounced when the task requires cross-disciplinary reasoning or iterative verification against external tools \cite{ghafarollahi2025sciagents, swanson2025virtual, su2025many, xia2025sr, hu2026multi}.

Despite this momentum, no existing multi-agent system has been tailored to the generation of physics-constrained constitutive neural networks: CSGA and GenCANN \cite{tacke2025constitutive, tacke2025automating} operate as essentially single-LLM pipelines and provide no checks on the physical admissibility of the resulting models. The present work closes this gap and makes three main contributions. We introduce the first multi-agent LLM architecture for constitutive model generation, in which a Creator proposes a model tailored to the data at hand and an Inspector critically audits each proposal against the physical constraints that any admissible hyperelastic model must satisfy: thermodynamic consistency, stress symmetry, objectivity, material symmetry, polyconvexity, growth, energy normalization, stress normalization, and non-negativity of the strain energy. Second, we show that this specialization is not merely a conceptual improvement but a measurable one. Across two LLM backbones and four benchmark datasets covering isotropic and transversely isotropic materials, adding the Inspector raises the share of exported models that pass numerical validation of all physical constraints from 90\% to 95\% for Claude Opus 4.7 and from 47\% to 60\% for Kimi K2.5. At the same time, the generated models reach accuracies on par with or even surpassing expert-designed models and generalize strongly to unseen loading paths. Third, we deliberately formulate the paradigm in a technique-agnostic way: although we demonstrate it on CANNs, the same Creator–Inspector loop can be instantiated around PANNs, CKANs, or symbolic laws with only minor changes to the prompts, providing a general template for physics-aware, LLM-driven constitutive modeling that scales automatically with future progress in both agentic design and LLM capability.

\section{Background} \label{sec_Background}

\subsection{Continuum mechanics preliminaries} \label{sec_Background_continuum_mechanics}

Constitutive models map a deformation state to a stress response, and the physical constraints considered later are formulated in exactly these quantities. We therefore fix the kinematic and stress measures, following the standard treatment of \citet{holzapfel2002nonlinear}.

Let $\mathbf{X}$ and $\mathbf{x}$ denote the position of a material point in the reference and the current configuration. The deformation gradient $\mathbf{F}$ captures the local kinematics, and the right Cauchy-Green deformation tensor $\mathbf{C}$ follows as
\begin{equation*}
\mathbf{F}=\frac{\partial\mathbf{x}}{\partial\mathbf{X}}, \qquad \mathbf{C}=\mathbf{F}^T\mathbf{F}.
\end{equation*}
The homogeneous loading scenarios considered here are parameterized by at most two scalars. In uniaxial tension and compression, this is the stretch $\lambda=l/l_0$ of current to reference length, a diagonal component of $\mathbf{F}$. In simple shear, it is the amount of shear $\gamma=u/h$, the lateral displacement over the specimen height, an off-diagonal component. In pure shear, the specimen is stretched by $\lambda$ in one in-plane direction and held in the other, and only the thickness contracts to preserve volume. The biaxial protocols stretch a thin sheet in both in-plane directions, parameterized by the two stretches $\lambda_1$ and $\lambda_2$.

The principal scalar invariants of $\mathbf{C}$ are
\begin{equation*}
I_1=\operatorname{tr}(\mathbf{C}), \qquad I_2=\tfrac{1}{2}\left[\operatorname{tr}(\mathbf{C})^2 - \operatorname{tr}(\mathbf{C}^2)\right], \qquad I_3=\det(\mathbf{C}).
\end{equation*}
For incompressible materials, volume preservation implies $I_3=1$. The response of an isotropic material then depends only on $I_1$ and $I_2$. For transversely isotropic materials, the preferred fiber direction $\mathbf{n}$ defines the structure tensor $\mathbf{N}=\mathbf{n}\otimes\mathbf{n}$ and two additional pseudo-invariants,
\begin{equation*}
I_4=\mathbf{C}:\mathbf{N}, \qquad I_5=\mathrm{cof}(\mathbf{C}):\mathbf{N}.
\end{equation*}
In our setting, the fiber direction is not prescribed but learned by the model from the data. Throughout this work, we represent the strain energy in this invariant basis. The equivalent representation in terms of principal stretches is not considered here.

A material is termed hyperelastic when its response derives from a scalar strain-energy density function $\Psi$. This assumption describes rubber-like materials and many soft biological tissues well and is the exclusive setting considered here. Constitutive modeling then amounts to specifying $\Psi$ as a function $f$ of the invariants — of $I_1$ and $I_2$ for isotropic materials, and additionally of $I_4$ and $I_5$ for transversely isotropic ones. The isochoric stress follows by differentiation, and incompressibility is enforced through the volumetric term $-p\mathbf{F}^{-T}$ with the Lagrange multiplier $p$ acting as a hydrostatic pressure,
\begin{equation*}
\Psi=f(I_1,I_2) \;\;\text{or}\;\; \Psi=f(I_1,I_2,I_4,I_5), \qquad \mathbf{P}=\frac{\partial\Psi}{\partial\mathbf{F}}-p\mathbf{F}^{-T}.
\end{equation*}
The first Piola--Kirchhoff stress $\mathbf{P}$, with components $P_{ij}$, measures force per unit reference area and is therefore work-conjugate to $\mathbf{F}$. Where stresses are reported per unit current area, as for the skin dataset, we use the Cauchy stress $\boldsymbol{\sigma}=\mathbf{P}\mathbf{F}^T$, with components $\sigma_{ij}$.

\subsection{Constitutive artificial neural networks (CANNs)} \label{sec_Background_CANNs}

\begin{figure}[h]
  \centering
  \includegraphics[width=\linewidth]{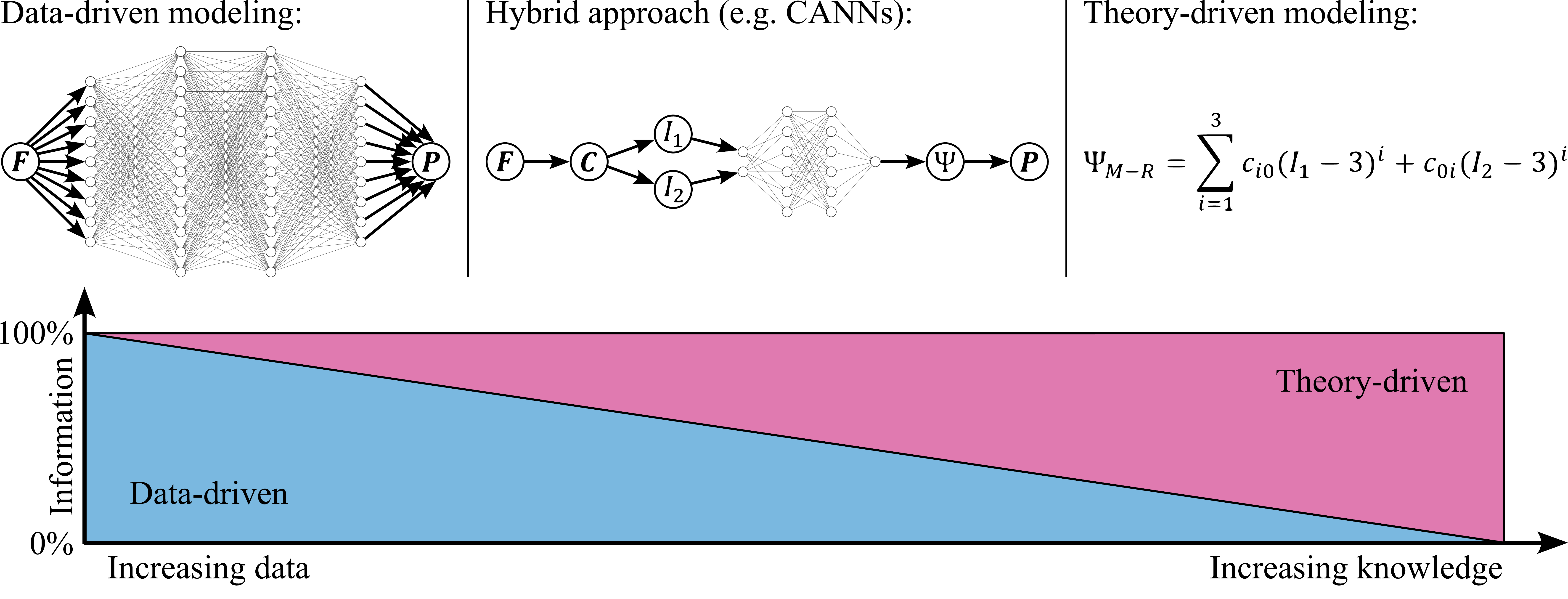}
  \caption{Spectrum of constitutive models for hyperelastic materials. Hand-crafted symbolic equations (right) embed physical structure by design, but their derivation is time-consuming. Pure black-box neural networks mapping strain directly to stress (left) are easy to train and highly flexible, but usually do not guarantee compliance with physical constraints. Hybrid approaches such as constitutive artificial neural networks (CANNs) sit in the middle: they represent the scalar strain-energy density $\Psi$ as a function of an invariant basis by a neural network and recover a stress measure like $\mathbf{P}$ by differentiation. This ensures compliance with certain concepts of physics while keeping the models easy and fast to obtain. Note that this scheme is simplified: all three approaches extend naturally, for example, to transverse isotropy.}
  \label{fig_architecture_cann_explanation}
\end{figure}

The central contribution of this work is a new paradigm for generating constitutive models through LLM-based agents, together with a mechanism to check that the generated models respect the physical constraints of hyperelasticity. The paradigm is technique-agnostic and, in principle, compatible with any parameterized family of constitutive models. For demonstration, we instantiate it with CANNs, which offer a favorable trade-off between flexibility and physical structure and are well established in the community.

As illustrated in Figure~\ref{fig_architecture_cann_explanation}, constitutive models for hyperelastic materials can broadly be placed on a spectrum ranging from hand-designed analytical laws, over hybrid formulations, to black-box models that map strain directly to stress. CANNs combine the physical reliability of hand-designed laws with the flexibility of black-box networks. They retain the invariant-based hyperelastic template of Section~\ref{sec_Background_continuum_mechanics}, in which a scalar strain-energy density $\Psi$ is expressed as a function of an invariant basis and a stress measure is recovered by differentiation, but represent the functional relationship between the invariants and $\Psi$ itself by a neural network. The architecture is chosen such that key physical requirements are satisfied by construction \citep{linka2021constitutive, linden2023neural}. Designing such an architecture for a new material, however, demands precisely the expertise whose necessity we aim to reduce. An LLM-generated CANN inherits the by-construction guarantees only if the LLM has designed it correctly, which is exactly what our framework sets out to verify. Since physics-augmented neural networks (PANNs) share the same philosophy, they could replace CANNs within our paradigm at no conceptual cost.

\subsection{Physical constraints and their validation} \label{sec_Background_constraints}

Hyperelastic constitutive models are subject to a well-established set of mathematical and physical conditions that any admissible model must satisfy. Following \citet{linden2023neural}, we consider nine such constraints: thermodynamic consistency, symmetry of the stress tensor, objectivity, material symmetry, polyconvexity, the volumetric growth condition, energy normalization, stress normalization, and non-negativity of the strain energy. The Creator is asked to enforce all of them by construction, the Inspector is asked to check that they are indeed satisfied, and our numerical tools provide an external check whose verdicts we use as ground-truth labels in Section~\ref{sec_Results}. The Inspector and the numerical validators are strictly separate components: the Inspector is an LLM agent that judges the generated code, whereas the validators are deterministic, sampling-based numerical procedures that probe each constraint at a finite but carefully chosen set of deformation states, rotations, and direction vectors. Apart from simple cases such as evaluating the energy and stress at the reference configuration, our validators therefore do not provide definite mathematical proofs: They declare a constraint as satisfied when no violation is found across all sampled states. The samples are designed to cover the relevant spaces densely enough that a violating model is, with high likelihood, exposed. Where a check depends on the material's symmetry class, we specify both an isotropic and a transversely isotropic variant. The detailed specifications are given in Appendix~\ref{sec_Appendix_validations}. The Inspector can invoke these validators as tools only in the dedicated tool-access variant introduced in Section~\ref{sec_Method_ablations}. Most of the constraints follow logically from a correctly implemented CANN architecture, so what must be checked is whether the generated implementation is indeed correct in this sense. 

\paragraph{Thermodynamic consistency}
A hyperelastic model must not create or destroy energy in purely elastic processes. The Clausius-Duhem inequality reduces, in this setting, to the requirement that the stress derives from a scalar strain energy potential $\Psi$ as $\mathbf{P_{\rm iso}}=\frac{\partial \Psi}{\partial \mathbf{F}}$ \citep{linden2023neural, gonzalez2019thermodynamically, upadhyay2019thermodynamics}. This potential structure is enforced by construction: a neural-network block predicts $\Psi$, the isochoric stress is obtained by automatic differentiation, and the volumetric contribution is added through a Lagrange-multiplier pressure $p$ that enforces incompressibility. Since implementation-level mistakes can still break it in practice, the numerical validator tests path independence directly along closed deformation loops and along open two-path comparisons. On closed loops, it checks that the cumulative work returns to zero relative to the total work expended and that the stress recovers its initial value upon loop closure, ruling out hidden statefulness. On open paths between the same end states, it checks that the two cumulative works agree with each other and that each of them reproduces the corresponding energy difference pointwise.

\paragraph{Symmetry of the stress tensor}
The balance of angular momentum requires the Cauchy stress $\boldsymbol{\sigma}$ to be symmetric, which in turn imposes a symmetry condition on the first Piola--Kirchhoff stress through $\boldsymbol{\sigma} = J^{-1}\mathbf{P}\mathbf{F}^T$ \citep{holzapfel2002nonlinear, linden2023neural}. This symmetry is automatically inherited when $\Psi$ is expressed as a function of invariants of $\mathbf{C}$ -- including, for anisotropic materials, the pseudo-invariants formed with the structure tensor -- rather than of the components of $\mathbf{F}$ or $\mathbf{C}$ directly. The validator computes $\boldsymbol{\sigma}$ from the model's stress prediction at a representative set of deformation gradients and checks that $\boldsymbol{\sigma}=\boldsymbol{\sigma}^T$ holds within a small tolerance.

\paragraph{Objectivity}
A constitutive model must not depend on the choice of observer: superimposing a rigid rotation on the current configuration must leave the energy unchanged, $\Psi(\mathbf{Q}\mathbf{F}) = \Psi(\mathbf{F})$ for all $\mathbf{Q}\in\mathcal{SO}(3)$ \citep{linden2023neural, holzapfel2002nonlinear}. Formulating $\Psi$ in terms of invariants of $\mathbf{C}=\mathbf{F}^T\mathbf{F}$ enforces this property by construction, independently of the material's symmetry class. The validator probes it numerically by comparing $\Psi(\mathbf{F})$ to $\Psi(\mathbf{Q}\mathbf{F})$ across a representative set of deformations and rotations.

\paragraph{Material symmetry}
The energy must also be invariant under the symmetry transformations of the material itself, $\Psi(\mathbf{F}\mathbf{Q}^T) = \Psi(\mathbf{F})$ for all $\mathbf{Q}$ in the symmetry group of the material \citep{linden2023neural, holzapfel2002nonlinear}. For isotropic incompressible materials the symmetry group is $\mathcal{O}(3)$, and the same invariant-based formulation that ensures objectivity also ensures material symmetry. The validator is structurally analogous to the objectivity check, but applies the rotation from the right and additionally includes reflections. For transversely isotropic materials the group is smaller, comprising only the rotations about the preferred fiber direction together with the reflections that map the fiber onto itself. Since the fiber orientation is learned by the model rather than prescribed, the validator first recovers it numerically from the model's own response and then verifies invariance under this reduced, fiber-aligned group.

\paragraph{Polyconvexity (and ellipticity)}
Polyconvexity of $\Psi$ is the standard route to material stability, which leads to favorable behavior in numerical applications: polyconvexity implies quasiconvexity, which in turn implies rank-one convexity, also known as ellipticity \citep{neff2015exponentiated, zee1983ordinary, linden2023neural}. Polyconvexity is much easier to enforce architecturally than ellipticity, which is why CANNs are typically built around it, and why we ask the Creator to enforce it by construction. For the numerical validation, we follow \citet{linden2023neural} and check the weaker but practically relevant ellipticity condition directly, namely that the rank-one form $(\mathbf{a}\otimes\mathbf{b}):(\partial^2\Psi/\partial\mathbf{F}\partial\mathbf{F}):(\mathbf{a}\otimes\mathbf{b})$ is non-negative for all direction pairs $(\mathbf{a},\mathbf{b})$ at all admissible deformations. The validator samples this form on a Fibonacci-distributed set of direction pairs and a set of admissible deformation gradients. For isotropic materials, objectivity and material symmetry allow the deformation to be reduced to diagonal form, so a grid of diagonal gradients suffices; for transversely isotropic materials the fiber cannot be rotated away, and the validator instead samples symmetric gradients whose principal axes are swept relative to the fiber, both within and out of the plane.

\paragraph{Growth condition}
The model should respond with infinite energy when the volume tends to zero or infinity, $\Psi\rightarrow\infty$ as $J\rightarrow 0^+$ or $J\rightarrow\infty$ \citep{linden2023neural, holzapfel2002nonlinear}. Under the incompressibility constraint $J=1$, only deformation states satisfying this constraint are admissible, so the growth condition is trivially satisfied. The validator returns success without any computation.

\paragraph{Energy normalization}
The strain energy must vanish in the reference configuration, $\Psi(\mathbf{F}=\mathbf{I})=0$ \citep{linden2023neural}. This is enforced architecturally by subtracting the network output evaluated at $\mathbf{C}=\mathbf{I}$ from the predicted energy. The validator simply evaluates the model at $\mathbf{F}=\mathbf{I}$ and checks that the returned energy lies within a small tolerance of zero.

\paragraph{Stress normalization}
In the reference configuration, the material must be stress-free, $\mathbf{P}(\mathbf{F}=\mathbf{I})=\mathbf{0}$ \citep{linden2023neural}. For isotropic incompressible materials this is automatically guaranteed by the Lagrange-multiplier pressure $p$ that enforces incompressibility. For transversely isotropic materials the scalar pressure cannot cancel the fiber-aligned part of the reference stress, so a stress-free reference must instead be enforced explicitly during construction. In either case, the validator evaluates the predicted stress at $\mathbf{F}=\mathbf{I}$ and verifies that all of its entries are below a small tolerance.

\paragraph{Non-negativity of the strain energy}
Finally, the strain energy must be non-negative for every admissible deformation, $\Psi(\mathbf{F})\geq 0$ \citep{linden2023neural}. For isotropic incompressible materials, this property is automatically inherited from the combination of polyconvexity, energy normalization, and stress normalization, because both isotropic invariants attain their minimum $I_1=I_2=3$ exactly at the reference configuration, so every monotonically increasing contribution is minimized there. For transversely isotropic materials, no such guarantee exists: the fiber pseudo-invariants can fall below their reference values under admissible deformations, and the non-negativity of the energy can in general only be verified numerically \citep{linden2023neural}. The validator therefore checks the condition explicitly in either case by evaluating $\Psi$ across the same representative set of deformation gradients used for the other checks and verifying that no negative value appears.

\section{Method} \label{sec_Method}

The central motivation of this work is to lower the barrier of accessibility to constitutive modeling. As the preceding sections and the extensive literature on the subject make clear \citep{linden2023neural, holzapfel2002nonlinear, neff2015exponentiated, linka2021constitutive, linka2023new}, developing an effective constitutive model is far from trivial: it requires a working knowledge of continuum mechanics, a careful choice of functional building blocks, and a practical implementation that respects a substantial list of physical constraints. Recent advances in large language models (LLMs) have shown impressive capabilities in scientific discovery and code generation at a level of usability that was unthinkable only a few years ago \citep{nejjar2025llms, wysocki2024llm}. It is therefore natural to ask whether they can also be leveraged to carry part of the constitutive modeling burden.

Our approach follows and extends a short but rapidly evolving line of research. The scientific generative agent (SGA) of \citet{ma2024llm} was the first to propose a general bilevel framework in which an LLM generates symbolic scientific hypotheses in the outer loop, while a differentiable simulator calibrates their continuous parameters in the inner loop. SGA was demonstrated on constitutive law discovery from motion observations and on molecular design, and consistently outperformed pure chain-of-thought or single-loop LLM baselines.

The constitutive scientific generative agent (CSGA) of \citet{tacke2025constitutive} specialized SGA for hyperelastic constitutive modeling. It injects domain knowledge into the prompt---such as isotropy, incompressibility, an invariant- or principal-stretch-based functional basis, and the requirement that the stress vanishes in the reference configuration---and replaces the expensive forward simulation with a direct regression on tabulated stress--strain data. CSGA clearly outperforms SGA on standard benchmarks but remains noticeably less accurate than purpose-built CANNs.

Most recently, \citet{tacke2025automating} closed this accuracy gap with GenCANN, in which the LLM no longer proposes a symbolic law but instead generates, on demand, a full CANN tailored to the material and dataset at hand. The generated networks match or exceed the accuracy of human-designed CANNs while retaining a plain-text interface. What GenCANN does not provide, however, is any systematic check that the generated networks satisfy the physical constraints listed in Section~\ref{sec_Background_constraints}. This is precisely the gap that we close in the present work.

\subsection{Multi-agent architecture: Creator and Inspector}

Rather than proposing yet another single-agent pipeline, we introduce a more general paradigm that we believe is a natural next step for LLM-driven constitutive modeling: two complementary agents that together pursue a common goal. A \emph{Creator} focuses on proposing an expressive and well-fitting constitutive model, while an \emph{Inspector} critically audits the Creator's proposal against the physical constraints discussed in Section~\ref{sec_Background_constraints}. This contrast between a generator and a critic is a classical divide-and-conquer strategy - it underlies the entire architecture of generative adversarial networks, and has recently been shown to be particularly effective in multi-agent LLM systems, where a clean separation between generating and inspecting agents consistently improves factual correctness and reasoning over single-model pipelines \citep{du2024improving, gridach2025agentic, chen2024autoagents, liu2024a}. Figure~\ref{fig_architecture_genncann_plus_inspector} illustrates this shift from the single-creator pipelines of earlier work to the Creator-Inspector paradigm adopted here.

Importantly, the paradigm itself is independent of the specific family of constitutive models at its core. The Creator could, in principle, generate symbolic laws, PANNs, CKANs, or any other parameterized constitutive model, and the Inspector would remain responsible for checking the same physical constraints. We deliberately speak of models rather than CANNs to emphasize this generality; our use of CANNs in the experiments that follow is a concrete and well-established demonstration vehicle, not a defining feature of the approach.

\subsection{Pipeline}

\begin{figure}[h]
  \centering
  \includegraphics[width=\linewidth]{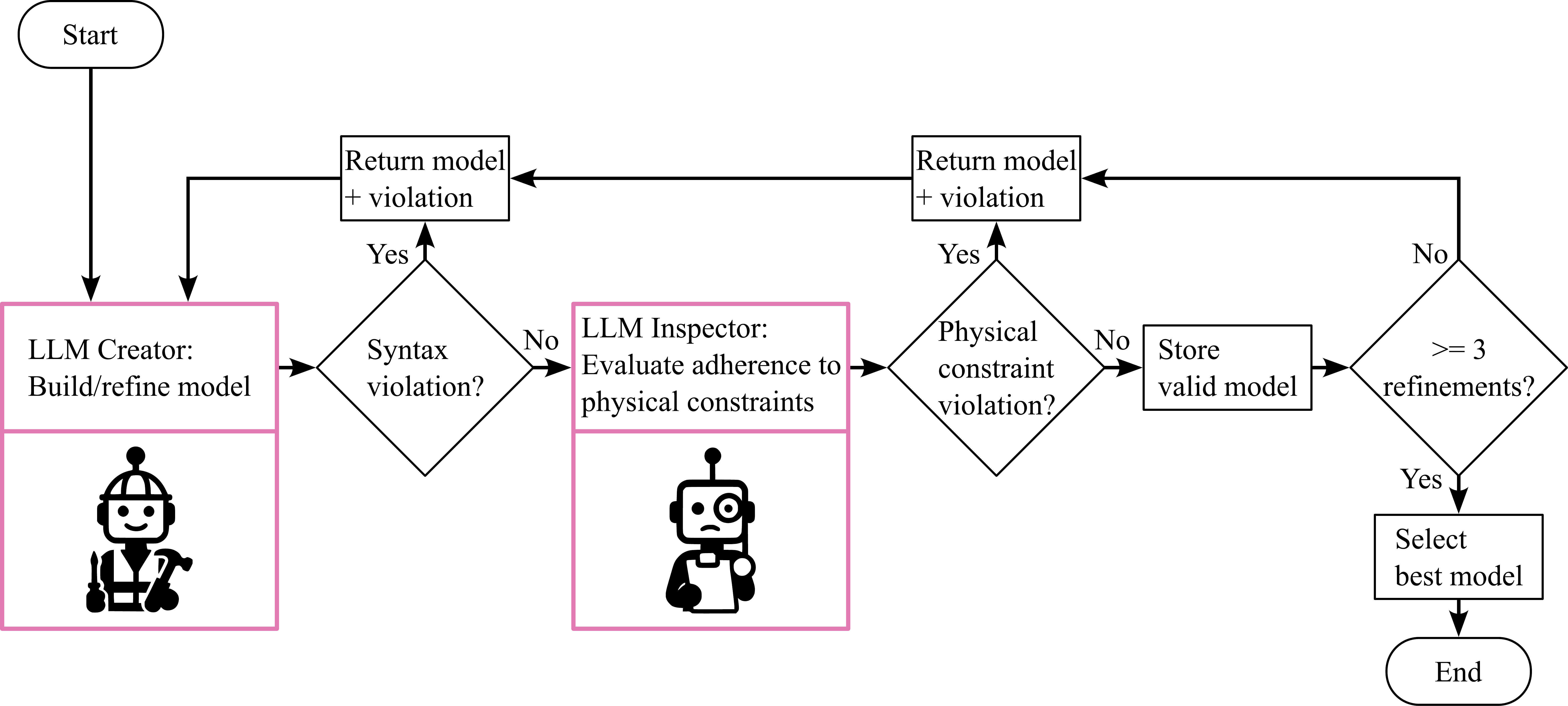}
  \caption{Detailed workflow of the Creator-Inspector pipeline. The Creator is prompted to implement a constitutive model and returns a Python implementation. The implementation is first subjected to automated syntactic checks (executability, training, prediction, save/load); failures are returned to the Creator for correction. Valid implementations are forwarded to the Inspector, which audits them against the nine physical constraints discussed in Section~\ref{sec_Background_constraints} and returns a verdict per constraint. Flagged violations are sent back to the Creator with the Inspector's justifications; fully approved models are exported. Once a valid model is obtained, the pair is invited to refine it twice based on training-data predictions and $R^2$/MSE metrics, so that each full run yields three inspector-approved models, of which the most accurate is selected.}
  \label{fig_architecture_detailed_flowchart}
\end{figure}

The complete workflow is illustrated in Figure~\ref{fig_architecture_detailed_flowchart} and proceeds as follows.

The Creator is prompted to implement a constitutive model for the given material class and dataset. The prompt briefly lists the nine physical constraints of Section~\ref{sec_Background_constraints}, without reproducing their full mathematical treatment, and requests a concrete Python implementation. The Creator responds with a complete model in its reply, from which the code is extracted automatically.

Before any inspection takes place, the proposed model is subjected to a set of automated syntactic checks. We verify that the Python code is executable, that the model can be instantiated, trained on the provided data, used for prediction, and correctly saved and loaded. Some failures are caused by outright syntax errors from the LLM; others by subtle incompatibilities between the LLM-generated model and the surrounding static code used for training, evaluation, and serialization. Whenever one of these checks fails, the model is returned to the Creator with a concise description of the failed check and a request for correction. Only implementations that pass all syntactic checks are forwarded to the Inspector.

The Inspector is then prompted with the full Python script of the model and with the same list of nine physical constraints, enriched by short practical hints on how each constraint can be enforced architecturally (for example, by using an input-convex subnetwork for polyconvexity or by deriving stresses from a scalar potential for thermodynamic consistency). The Inspector is required to return its verdict stating, for every constraint, whether it is fulfilled or violated, together with a short justification.

If the Inspector flags at least one constraint as violated, the model is sent back to the Creator together with the list of flagged constraints and the Inspector's justifications. The Creator is then asked to modify its implementation so as to address these concerns. If all constraints are flagged as fulfilled, the implementation is stored and exported as a valid model.

Once a valid model is obtained, the Creator-Inspector pair is asked to refine it. The current implementation is passed back to the Creator, together with a tabular summary of its predictions on the training data and the corresponding $R^2$ and mean squared error (MSE) values, and the Creator is invited to propose an improved variant. Each refined model again passes through the syntactic checks and the Inspector before being exported. The refinement loop is executed twice after the initial generation, so that each full run produces, in total, three inspector-approved models. The most accurate one is then selected as the output of the run.

\subsection{Two natural ablations} \label{sec_Method_ablations}

Two questions arise naturally from this design, and we address both by means of dedicated ablations.

The first question concerns the value of specialization itself. Since GenCANN \citep{tacke2025automating} already demonstrated that an LLM can generate accurate constitutive models in a largely single-agent fashion, it is not self-evident that an additional Inspector is worth its computational and conceptual cost. To probe this, we run a full ablation in which only the Creator is active and every syntactically valid model is exported without further verification. Because our prompts differ slightly from those of GenCANN, this ablation also serves as a fair baseline for the Inspector's added value within our setup.

The second question concerns the Inspector's own limitations. In principle, any physical constraint could simply be checked by directly running the numerical validation tools of Section~\ref{sec_Background_constraints}. Doing so unconditionally would, however, conflate the capabilities of the LLM with those of the validators and obscure what the LLMs themselves actually understand about hyperelastic admissibility. We therefore test a second variant in which the Inspector is additionally granted access to the numerical validation tools. All prompts remain unchanged, except for a single sentence informing the Inspector that these tools are available and instructing it to invoke them only when it is genuinely uncertain about a specific constraint. This setup probes whether the LLM can accurately estimate its own uncertainty and use external tools selectively and effectively, rather than as a substitute for its own reasoning.

Beyond these two ablations, we additionally repeated all configurations on the synthetic rubber dataset with minimal prompts that merely name the physical constraints without explaining them. This study probes how much of the system's reliability rests on the domain knowledge provided in the prompts and is reported in Appendix~\ref{sec_Appendix_minimal_prompts}.

\subsection{LLM backbones}

We instantiate the Creator and the Inspector with two LLM backbones and run all experiments for each of them. As a frontier proprietary model we use Anthropic's Claude Opus 4.7, and as an open-source counterpart we use MoonshotAI's Kimi K2.5. Comparing the two allows us to assess how robust our findings are with respect to the choice of backbone, and to what extent the observed behaviors are specific to a particular model family.

\subsection{Benchmark problems} \label{sec_Method_problems}

To keep the focus on the LLM-powered agentic setup itself, we restrict our benchmark problems to hyperelastic materials, all of which we model as incompressible. We consider two isotropic materials and one transversely isotropic material, the latter to evaluate whether our approach also handles materials with a preferred fiber direction. Within the isotropic class, we select two materials represented by three datasets that are well established in the constitutive modeling literature and that together serve complementary purposes: a brain-tissue dataset that is particularly challenging to fit, a synthetic rubber dataset that permits the computation of ground-truth stresses for arbitrary deformations and thereby a clean assessment of generalization and extrapolation, and an experimental rubber dataset that anchors the synthetic one in physical reality. As the transversely isotropic material, we select porcine skin, a soft tissue whose collagen fibers are predominantly aligned along one direction, so that the tissue responds markedly stiffer along the fibers than transverse to them.

The brain-tissue dataset captures the mechanical response of human cerebral tissue \citep{budday2017mechanical, budday2017rheological, budday2019fifty}, which is known to be soft, nearly incompressible, strain-stiffening, and asymmetric in tension and compression. Accurate models of it are relevant for impact simulation, injury prediction, and protective design. We use the data reported by \citet{budday2017mechanical}, who performed mechanical tests on specimens excised from ten post-mortem human brains within 60 hours of death. Of the anatomical regions sampled, we consider the cortical gray matter. Three loading modes were applied, namely uniaxial tension, uniaxial compression, and simple shear, each in loading--unloading cycles, and the mean stress across the hysteresis loop was retained as the effective elastic response. Only 17 stress-strain points are reported per loading mode, which poses a genuine challenge for models expected to generalize well. Such sparsity is, however, characteristic of soft-tissue experiments and reflects standard practice in the field.

The second isotropic material is rubber, the prototypical hyperelastic solid, which sustains large and fully reversible strains far beyond the regime of linear elasticity and whose accurate modeling underpins the design of engineering components such as tires and seals. We rely on two complementary sources: the classic experimental dataset of \citet{treloar1944stress} and a synthetic counterpart introduced alongside the original CANN paper \citep{linka2021constitutive}, which represents a similar but fictitious rubber-like material. Both cover the same three loading protocols, namely uniaxial tension, equibiaxial tension, and pure shear, with 15 samples per protocol, and we adopt the train-test split of the original CANN publication. Since the underlying material of the synthetic dataset is known in closed form, exact stresses can be computed for arbitrary deformations, which enables a controlled evaluation of how well a trained model generalizes beyond the loading paths seen during training.

The porcine skin dataset consists of publicly available biaxial experiments on skin specimens \citep{tac2022datadrivenmodeling, tac2022datadriventissue} with a total of 402 stress--stretch points. Biaxial testing is essential for this material: uniaxial data suffice to calibrate isotropic models, but the directional information required to identify fiber-induced anisotropy only becomes visible when the specimen is loaded in two directions simultaneously. The dataset covers five loading protocols: equibiaxial tension, strip-axial tension in each of the two in-plane directions, and off-axial tension biased toward either direction. Since the stress is recorded in both in-plane directions, each protocol contributes two stress--stretch curves, yielding ten in total. We hold out a random 20\% of the data points as a test split. As the baseline for this dataset, we adopt the CANN identified through a systematic hyperparameter search by \citet{linka2023automatedmodel}, which, in contrast to our setup, was trained on all available data points without a test split.

\section{Results} \label{sec_Results}

We organize the presentation of our results around a set of research questions that arise naturally from the methodology introduced above.

\subsection{Does the specialization into a multi-agent system bring any advantage over the Creator alone?}\label{sec_Results_agents}

\begin{figure}[h]
  \centering
  \includegraphics[width=\linewidth]{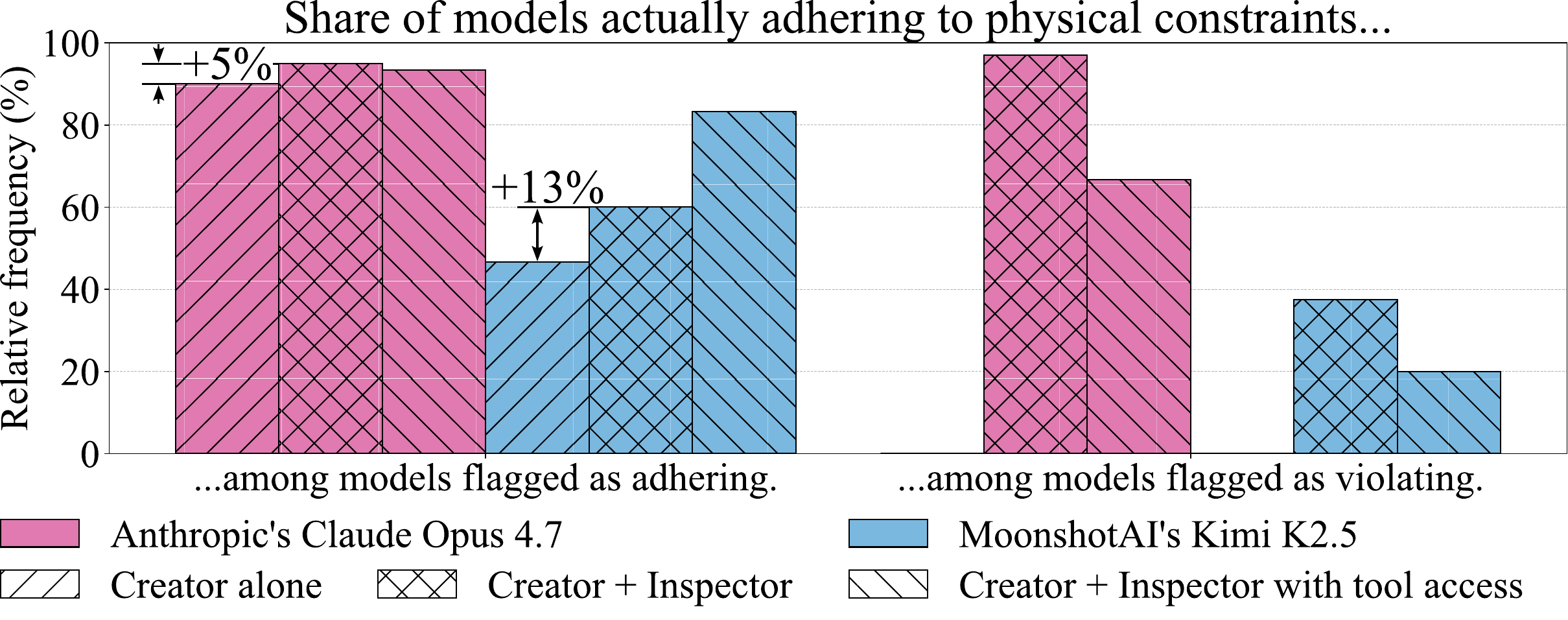}
  \caption{Adherence of generated models to physical constraints, split by Inspector verdict. For each configuration — Creator alone, Creator with Inspector, and Creator with tool-enabled Inspector, each powered by Claude Opus 4.7 or Kimi K2.5 — we report the fraction of models that truly adhere among those the Inspector flagged as adhering (left) and as violating (right). Since the Creator-alone setup has no Inspector, all its models are implicitly counted as flagged adhering. True adherence is established via the numerical validation procedure described in Section~\ref{sec_Background_constraints}.}
  \label{fig_predictions_cann_verifications_pct}
\end{figure}

The central claim of this work is that splitting the task of constitutive model generation between two specialized agents — the Creator and the Inspector — enables automatic and critical validation of the proposed models, and in doing so makes the overall system more trustworthy than a single LLM acting on its own. Whether this claim holds translates into a concrete question: how often does each approach actually produce models that respect the underlying physical constraints?

To answer this, we need a reliable notion of ground truth. We obtain it by numerically validating every proposed model using the procedure described in Section \ref{sec_Background_constraints}, and we consider a model to truly adhere to the physical constraints if and only if it passes these checks. This must not be confused with the Inspector's verdict, which only determines how a model is flagged: models judged acceptable by the Inspector are exported as the outcome of the refinement round and labeled flagged as adhering, while from those flagged as violating we export a random subset so that we can also assess how accurate the Inspector's negative verdicts are.

Pooled across 10 runs on four benchmark problems with three refinement rounds per run, Figure \ref{fig_predictions_cann_verifications_pct} reports the relative frequency of models that truly adhere among those the Inspector flagged as adhering (left) and as violating (right). From a practical standpoint, the most informative quantity is the share of flagged-as-adhering models that truly adhere, as it directly reflects how much a user can trust a model returned by the system. For Claude Opus 4.7, the Creator alone reaches 90\%, and adding the Inspector lifts this to 95\%. For Kimi K2.5, the corresponding numbers are 47\% without and 60\% with the Inspector. Two observations stand out. First, even between frontier LLMs the absolute performance differs substantially, and with Claude Opus 4.7 the proposed approach becomes highly reliable for this problem setting, with 19 out of 20 exported models withstanding all numerical constraint checks. Second — and this is the answer to the question posed in this subsection — the gains contributed by the Inspector, 5 and 13 percentage points, are consistent across both backbones and show that the second, specialized agent meaningfully increases our confidence in the physical validity of the generated models.

\subsection{How reliable is the Inspector's verdict?}\label{sec_Results_verification}

So far we have compared the Creator alone against the Creator–Inspector pair. As described in Section \ref{sec_Method_ablations}, we additionally evaluated a third variant in which the Inspector has access to the numerical validation tools used to check adherence to the physical constraints.

Since these tools also supply the ground-truth labels of our analysis, their reliability deserves a brief comment. As discussed in Section \ref{sec_Background_constraints}, they are sampling-based procedures rather than mathematical proofs. For the present task, this is less limiting than it may appear: the validators do not certify arbitrary black-box functions but probe implementations of an architecture that, implemented correctly, satisfies the constraints by construction — and when an LLM fails at this, the smoothness of the underlying strain-energy functions lets the violation extend over whole regions of deformation space rather than isolated points, so that a dense sample is unlikely to miss it. In addition, the sampled deformation states reach far beyond both the training data and the loading regimes of typical applications. What remains is an asymmetry between the two verdicts: a detected violation constitutes a counterexample and is therefore definitive, whereas a pass records only that no counterexample was found. The adherence rates reported here can thus only err on the generous side.

\begin{figure}[h]
  \centering
  \includegraphics[width=\linewidth]{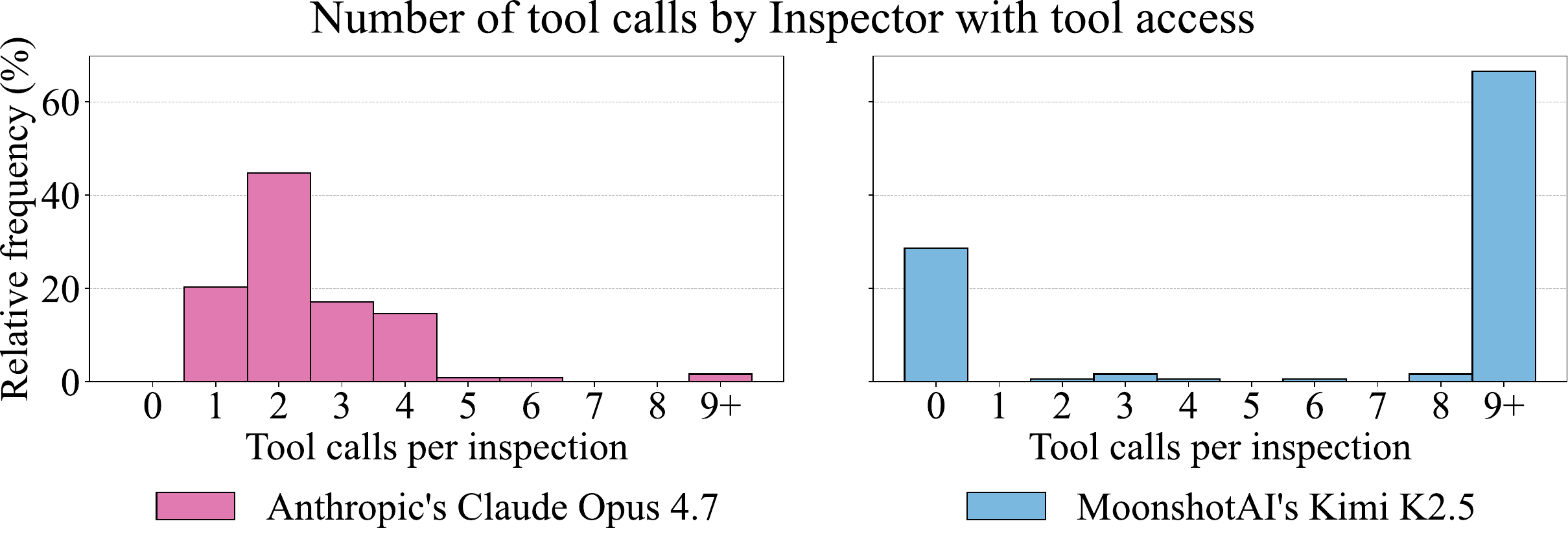}
  \caption{Tool usage of the Inspector per inspection. Distribution of the number of tools called by the Inspector in a single inspection, shown separately for Claude Opus 4.7 and Kimi K2.5. The Inspector is instructed to invoke tools only when uncertain. Opus follows this instruction closely and predominantly calls just one or two tools, whereas Kimi exhibits a bimodal pattern, either calling all available tools or none at all.}
  \label{fig_predictions_n_tool_calls}
\end{figure}

Figure \ref{fig_predictions_cann_verifications_pct} reports the relative frequency of truly adhering models among those flagged as adhering (left) and as violating (right), for all three configurations and for both Claude Opus 4.7 and Kimi K2.5. Looking at the flagged-as-adhering side first, one would expect the share of truly adhering models to grow as the pipeline becomes more elaborate, and especially when the Inspector is granted tool access. For Kimi this expectation is met: the share rises from 47\% (Creator alone) to 60\% (with Inspector) and further to 83\% with tool access. For Opus, however, the trend breaks. The Inspector without tools already lifts the Creator from 90\% to 95\%, leaving little room for tool access to improve further — and in fact the share even drops slightly to 93\%.

The flagged-as-violating side reveals a second, equally striking pattern. Of the models that the Opus Inspector (without tools) flags as violating, 97\% in fact adhere to the physical constraints. In other words, when in doubt, Opus tends to err on the side of raising a flag — a conservative and, for our purposes, desirable behavior. With tool access, these unfounded flags become far rarer. The Kimi Inspector behaves differently: the majority of the models it flags as violating do violate — 63\% without and 80\% with tool access — so it tends to raise a flag only when it clearly sees a problem.

Figure \ref{fig_predictions_n_tool_calls} helps explain these patterns by showing how many tools the Inspector calls per inspection when tool access is available. The Inspector is instructed to invoke tools only when it is unsure, since we want to measure the LLMs' own capabilities in creating and inspecting models rather than those of the numerical validators. Opus follows this instruction closely, most often calling just one or two tools to resolve specific doubts. Kimi, in contrast, exhibits a bimodal behavior: it either calls all available tools or none at all, with almost nothing in between.

Taken together, these observations paint a coherent picture of the two backbones. Kimi appears inherently submissive: without tools it readily accepts violating models, and it tends to flag a model as violating only when it is very certain or when the tools explicitly confirm a problem. This explains both the low raw performance without tools and the high precision on the violation side. Opus, on the other hand, is naturally more critical, which is why the Inspector already delivers excellent performance without tools. Its slight drop in performance under tool access is consistent with its sparse tool usage: Opus respects the "only-when-unsure" instruction, but in a handful of cases it is incorrectly confident about specific constraints and then does not consult the tools.
Conversely, it sometimes doubts constraints that are fulfilled and then draws false confidence from the tools it calls to check them. We return to this point in Section \ref{sec_Results_types}.

\subsection{Are the violations raised by the Inspector solved by the Creator?}\label{sec_Results_transitions}

\begin{figure}[h]
  \centering
  \includegraphics[width=\linewidth]{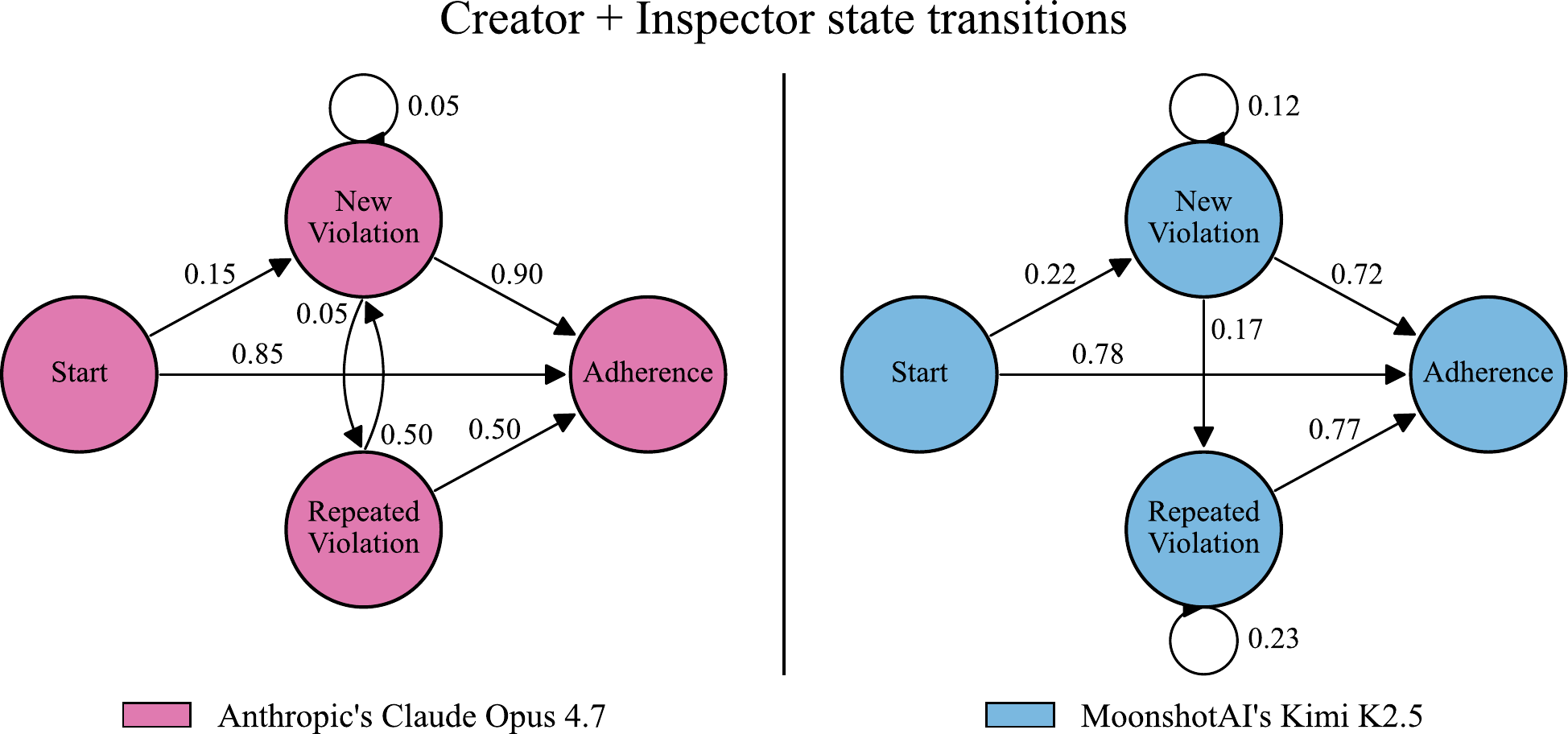}
  \caption{State transitions between Creator and Inspector across refinement rounds. Relative frequencies of transitions between the states Start, New violation, Repeated violation, and Adherence, shown separately for the pipelines powered by Claude Opus 4.7 and Kimi K2.5. States are defined solely by the Inspector's verdict. Runs with and without tool access are pooled.}
  \label{fig_predictions_markov_chain}
\end{figure}

To assess how effectively the two agents collaborate, we track the states that a Creator–Inspector pair can be in over the course of a refinement loop and report the relative frequencies of transitions between these states in Figure \ref{fig_predictions_markov_chain}. We define four states: Start, New violation, Repeated violation, and Adherence. Since we are interested in the agents' own behavior, the state is determined entirely by the Inspector's verdict, not by the numerical ground truth: a violation is a violation flagged by the Inspector, and adherence is adherence flagged by the Inspector. For this analysis we pool the runs with and without tool access into a single Inspector category.

Effective collaboration would show itself in two ways: most models should reach adherence quickly, and when violations are raised they should rarely persist across iterations. For the Opus-powered pair, both criteria are clearly met. Already at the start, 85\% of the Creator's proposals are flagged as adhering to all physical constraints, leaving only 15\% in the violation state. Of these, 90\% are resolved directly in the next iteration, while 5\% persist as repeated violations and 5\% lead to new ones. As repeated violations thus occur in only 5\% of the initial 15\%, they are clear outliers. The Kimi-powered pair shows similar characteristics, but performs consistently worse at every stage. It starts from a lower baseline — 78\% adherence and 22\% violations after the first proposal — and resolves new violations directly in only 72\% of the cases (compared to 90\% for Opus), while 17\% persist as repeated violations and 12\% turn into new ones. Once a repeated violation is reached, most cases still end well: 77\% are resolved into full adherence, and the remaining 23\% stay repeated violations for another iteration.

Overall, the Opus-powered Creator–Inspector pair collaborates effectively: violations raised by the Inspector are resolved in 90\% of the cases, and hardly any persist beyond a single additional iteration. The Kimi-powered pair collaborates noticeably less effectively: it shows a similar qualitative behavior, but consistently weaker numbers — more violations at the start, a lower direct resolution rate of 72\%, and more rounds spent on repeated violations.

\begin{figure}[!h]
  \centering
  \includegraphics[width=\linewidth]{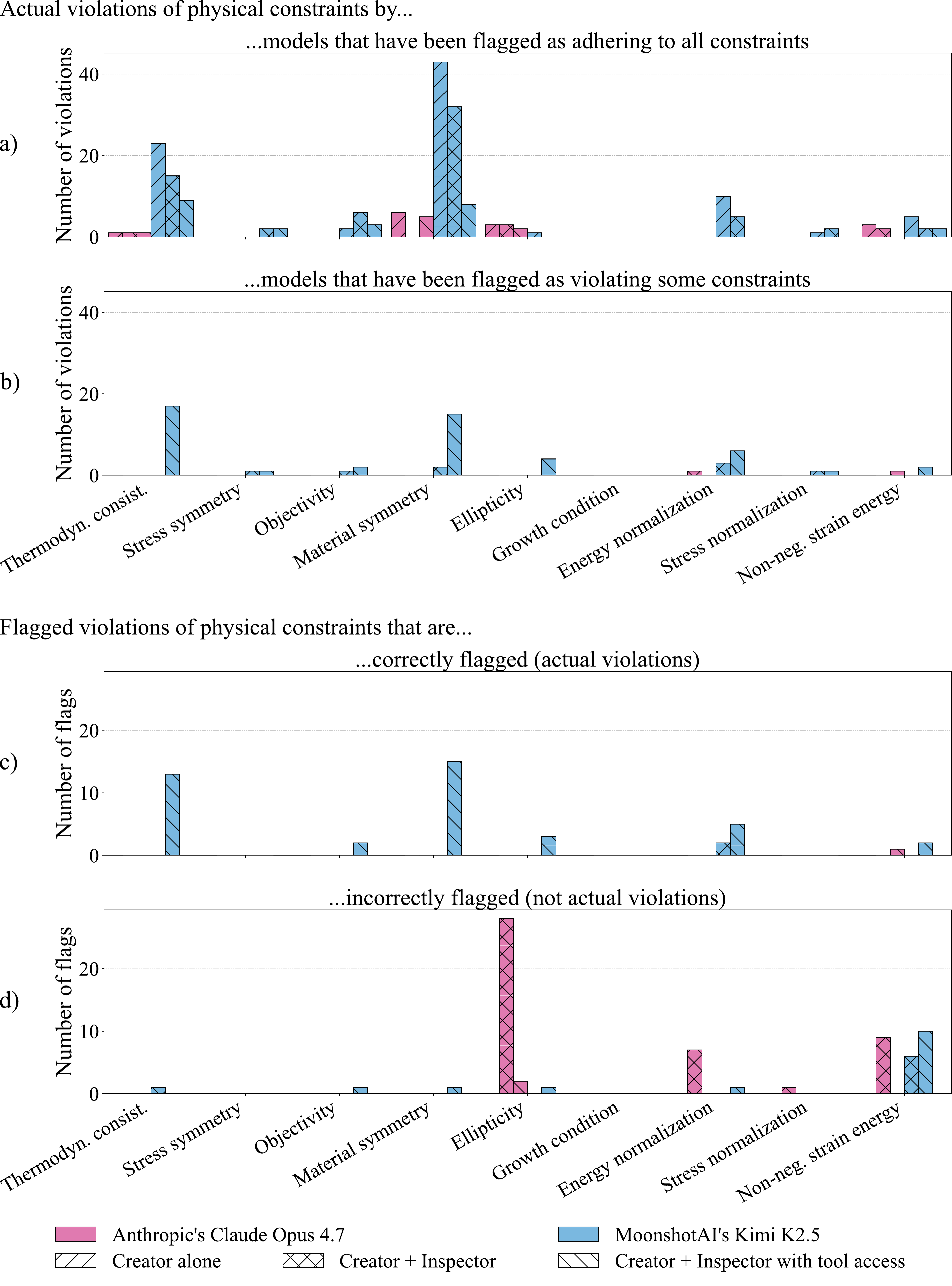}
  \caption{Distribution of physical constraint violations by type, separated into actual vs. flagged and correct vs. incorrect.}
  \label{fig_predictions_constraint_types}
\end{figure}

\subsection{Zooming in: Which violations actually occur, which are detected, and which are missed?}\label{sec_Results_types}

Figure \ref{fig_predictions_constraint_types} breaks violations of the physical constraints down by type across four charts. Charts a) and b) show actual violations, as confirmed by the numerical validation of Section \ref{sec_Background_constraints}: Chart a) for models that the Inspector flagged as adhering to all constraints, and Chart b) for models that the Inspector flagged as violating some constraints. Charts c) and d) show flagged violations raised by the Inspector: Chart c) those that our numerical validation confirms to be correct, and Chart d) those that it does not. We use the same y-axis range in Charts a) and b) for direct comparability. The overall counts in b) are much lower than in a), because every refinement round exports exactly one Inspector-approved model, whereas violating models are only exported as a random subset for analysis; this is an artifact of our experimental design and does not affect the relative distributions across constraint types, which is what Figure \ref{fig_predictions_constraint_types} is about.

We first look at the actual violations in Charts a) and b). Among models flagged as adhering, the Opus-powered configurations produce very few violations overall, and the Opus-powered Creator–Inspector pair without tools produces hardly any — most often concerning ellipticity — consistent with the 95\% adherence reported in Section \ref{sec_Results_agents}. The violations that slip through the tool-enabled variant stem mostly from material symmetry, and those of the Creator alone concern mainly material symmetry, ellipticity, and non-negativity of the strain energy. The Kimi-powered configurations show a broader pattern dominated by material symmetry, followed by thermodynamic consistency and energy normalization, with the overall count decreasing from the Creator alone to the Creator–Inspector pair and further to the tool-enabled variant. Among models flagged as violating, the Opus-powered configurations contribute almost nothing: the Creator alone implicitly flags every model as adhering, and the flagged-as-violating models of the Creator–Inspector pair mostly turned out to truly adhere (cf.\ Section \ref{sec_Results_verification}) — of those exported for analysis, a single one indeed violates energy normalization without tool access, and a single one violates non-negativity of the strain energy with it. The Kimi distribution, in contrast, closely mirrors the one in Chart a), indicating that the constraints Kimi's Creator struggles with are the same regardless of whether the Inspector later catches them.

The flagged violations in Charts c) and d) reflect the two characteristic behaviors identified in Section \ref{sec_Results_verification}. Correctly flagged violations come almost exclusively from the tool-enabled Kimi Inspector, which is less a particular strength than a consequence of its bimodal tool usage: when it calls tools, it calls all of them, so most of the flagged violations have effectively been pre-confirmed by the same numerical checks we use as ground truth. Incorrectly flagged violations come largely from the Opus Inspector without tools, whose uncertainty is concentrated on ellipticity and, to a lesser extent, on energy normalization and non-negativity of strain energy; with tool access these doubts are resolved numerically and the unnecessary flags largely disappear. The Kimi Inspectors contribute a further 21 incorrect flags, most of them concerning the non-negativity of the strain energy.

Two takeaways stand out. For Opus, when in doubt — especially about ellipticity, energy normalization, and non-negativity of strain energy — the Inspector prefers to flag rather than to accept, which is inefficient but aligned with our goal of erring on the safe side. For Kimi, the violations that slip through are not caused by a blind spot for any single constraint but by small inaccuracies spread across the same constraints that the Kimi-powered Creator already finds hardest to satisfy.

\begin{figure}[h!]
  \centering
  \includegraphics[width=\linewidth]{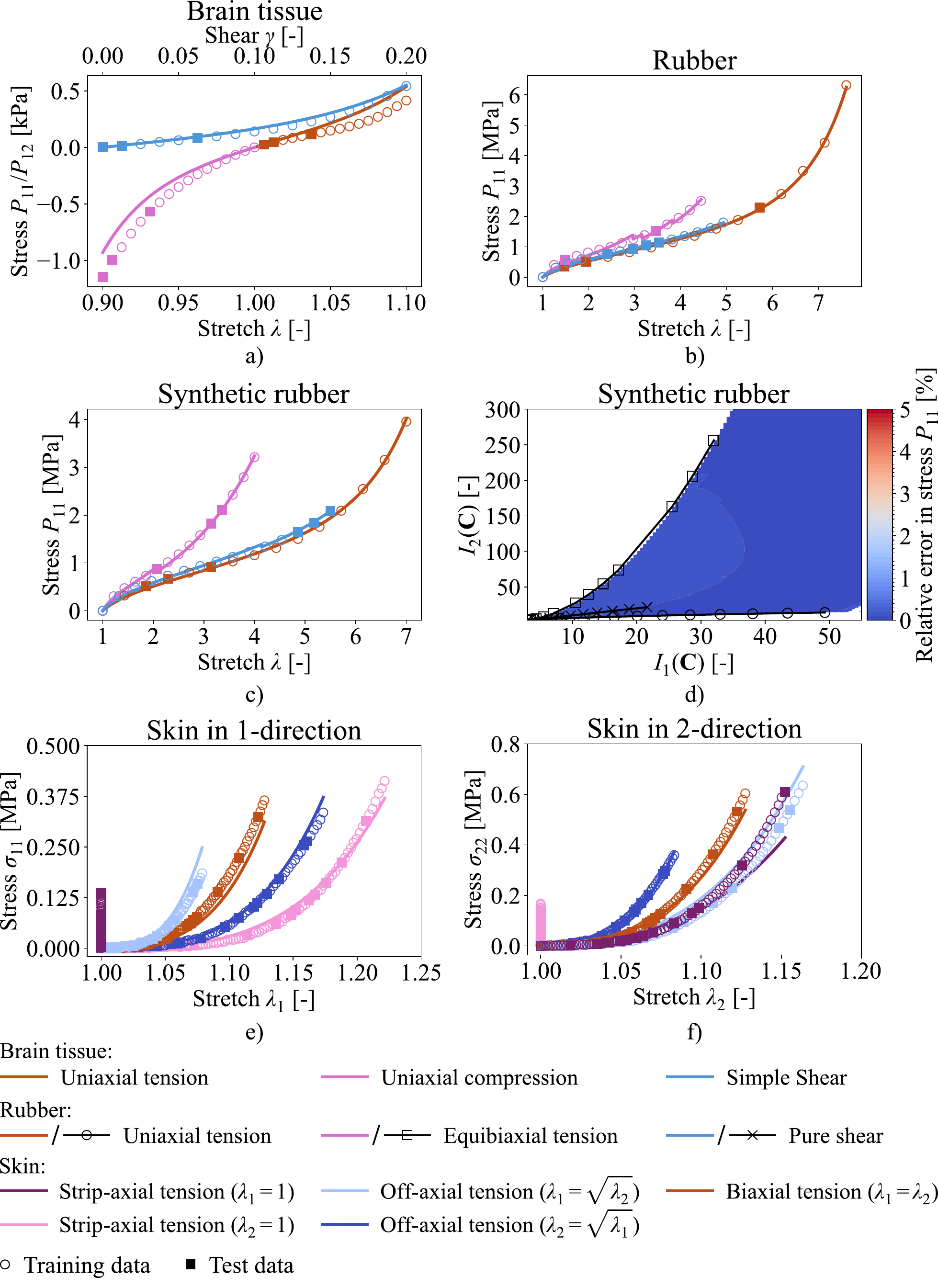}
  \caption{Models generated with Claude Opus 4.7 with Inspector without tool access. For each dataset, the most accurate adherent model out of ten runs is shown.}
  \label{fig_predictions_curves_opus}
\end{figure}
\begin{figure}[h!]
  \includegraphics[width=\linewidth]{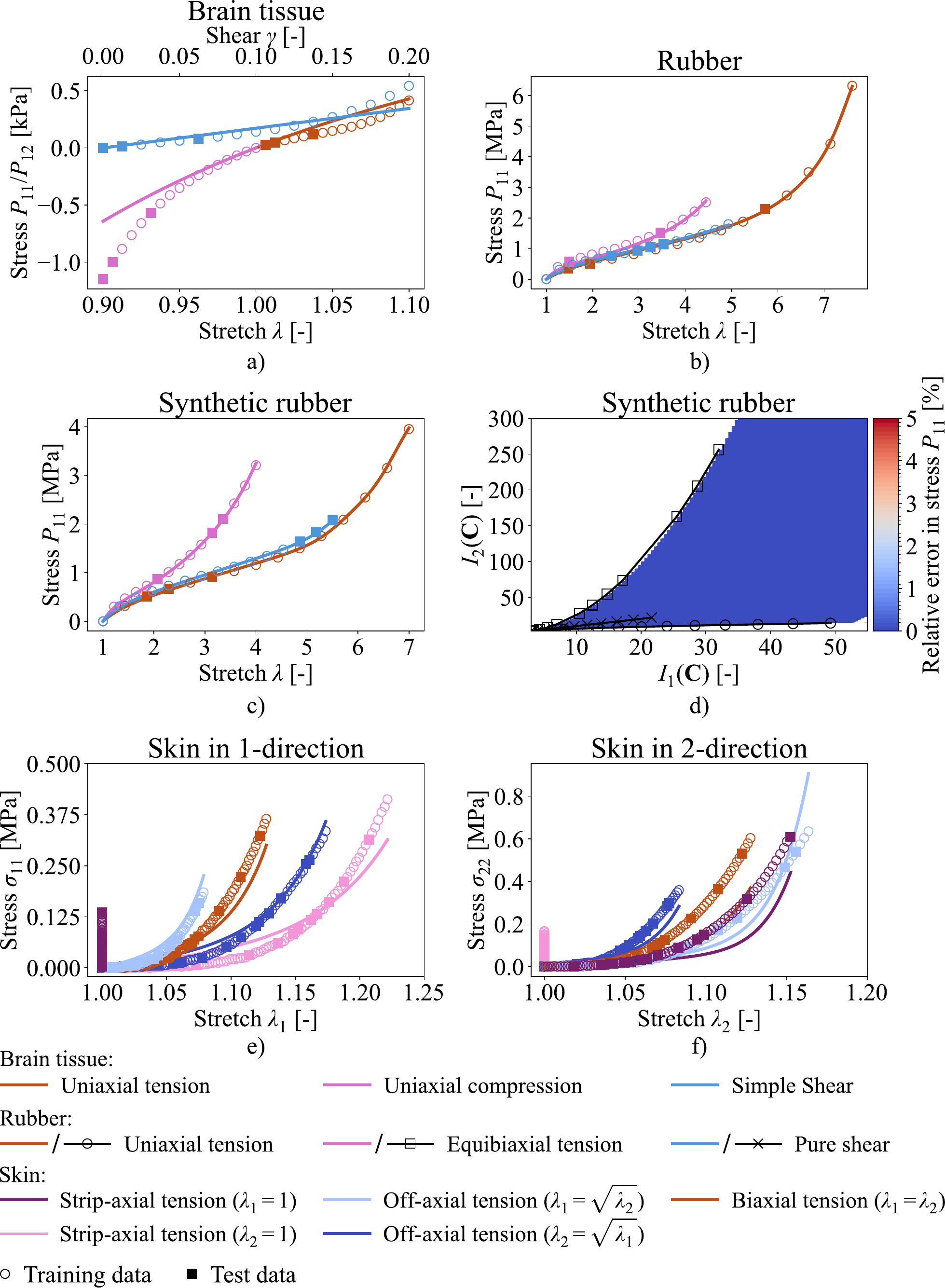}
  \caption{Models generated with Kimi K2.5 with Inspector without tool access. For each dataset, the most accurate adherent model out of ten runs is shown.}
  \label{fig_predictions_curves_kimi}
\end{figure}
\begin{figure}[h!]
  \includegraphics[width=\linewidth]{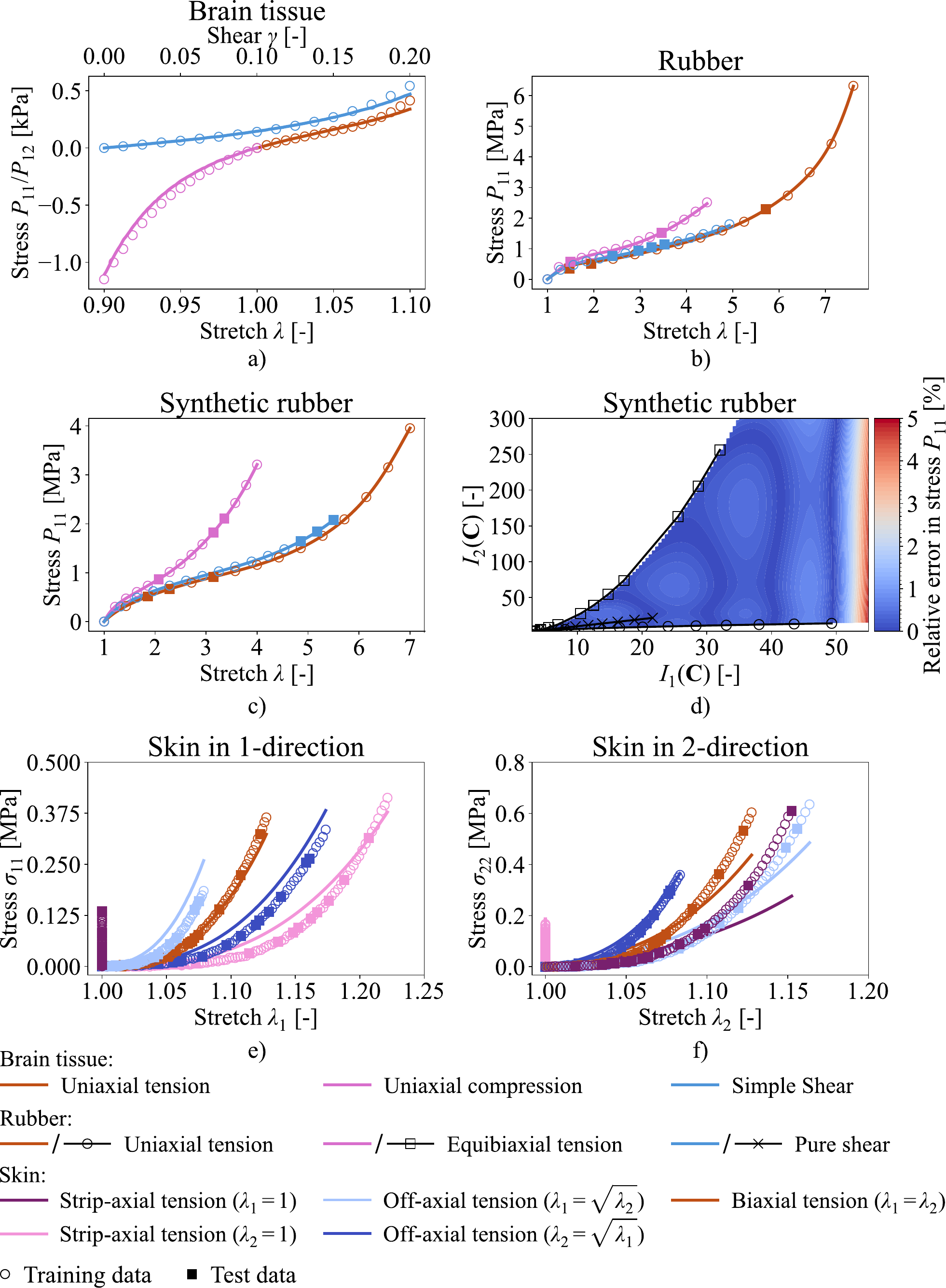}
  \caption{Models designed by human experts that serve as baselines \cite{pierre2023principal, linka2021constitutive, linka2023automatedmodel}.}
  \label{fig_predictions_curves_baseline}
\end{figure}

\subsection{How do the predictions of the generated models look in practice?}\label{sec_Results_predictions}

To get an impression of how the generated models behave in practice, we compare the predictions of models produced by our pipeline with established baselines from the literature. We focus on the setting with the Creator and Inspector but without tool access, and, from ten independent runs, select the most accurate model that also adheres to the physical constraints. This selection procedure reflects how we expect the pipeline to be used in practice: since it is built around stochastic LLMs, multiple runs should be performed and the best-performing model kept. To give the reader a concrete impression of the artifacts produced by the pipeline, the Python implementations of the two models selected for the experimental rubber dataset are shown in Appendix~\ref{sec_Appendix_exemplary_canns}.

Figure \ref{fig_predictions_curves_opus} shows the predictions for models generated with Claude Opus 4.7, Figure \ref{fig_predictions_curves_kimi} those generated with Kimi K2.5, and Figure \ref{fig_predictions_curves_baseline} the literature baselines. For brain tissue, we take the most accurate CANN reported across several optimization studies on this dataset \cite{linka2023automated, mcculloch2024sparse, pierre2023principal, budday2019fifty} as baseline \cite{pierre2023principal}. For both rubber datasets — the experimental and the synthetic one — we use the most accurate CANN from the original publication \cite{linka2021constitutive}. For porcine skin, we use the CANN identified through a systematic hyperparameter search by Linka et al.\ \cite{linka2023automatedmodel}. Charts a), b), c), e), and f) of Figures \ref{fig_predictions_curves_opus} - \ref{fig_predictions_curves_baseline} show the predictions on the training loading conditions: uniaxial tension, uniaxial compression, and simple shear for the brain data, and uniaxial tension, equibiaxial tension, and pure shear for the rubber data. For the skin data, the five biaxial protocols yield stresses in both in-plane directions, which we show separately: Chart e) reports the stress along the 1-direction and Chart f) the stress along the 2-direction.

On both rubber datasets, the baselines achieve near-perfect predictions, confirming that highly accurate constitutive models are well established for these commonly used benchmarks. Our generated models match this performance, reaching near-perfect accuracy across all loading conditions.

On the brain tissue data, the baseline — the most accurate model from several dedicated optimization studies \cite{linka2023automated, mcculloch2024sparse, pierre2023principal, budday2019fifty} — achieves $R^2$ values of 0.96 on tension, 0.99 on compression, and 1.00 on shear. It builds on principal stretches \cite{pierre2023principal}, which are known to capture the pronounced tension--compression asymmetry of brain tissue considerably better than invariants \cite{pierre2023principal, linka2023automated, mcculloch2024sparse, kuhl2024itoo, abdolazizi2025constitutive}. To keep our experiments focused, our setup prescribes invariants as the functional basis for all materials. Comparisons across functional bases can be found, for example, in the work of Abdolazizi et al.\ \cite{abdolazizi2025constitutive}. Within this basis we chose, the Kimi-powered pair reaches 0.614 on tension, 0.651 on compression, and 0.829 on shear, and the Opus-powered pair 0.453, 0.876, and 0.981. Measured against the expert-designed invariant-based models reported for this dataset \cite{linka2023automated, mcculloch2024sparse, abdolazizi2025constitutive}, our generated models are competitive.

The porcine skin data prove challenging even for the expert-designed baseline, which reaches a mean $R^2$ of 0.83 across the ten stress--stretch curves. The Opus-generated model surpasses this value with a mean of 0.90, while the Kimi-generated model stays slightly below it at 0.80. Interestingly, all three models share the same weak spot: the lowest scores occur for the stress component in the direction that is barely stretched, most prominently in the strip-axial protocols.

\subsection{Are the generated models capable of extrapolating beyond training data and generalizing beyond training loading paths?}\label{sec_Results_extrapolation}

To answer this question, we use Chart d) of Figures \ref{fig_predictions_curves_opus}, \ref{fig_predictions_curves_kimi}, and \ref{fig_predictions_curves_baseline}, which evaluate the models on loading scenarios that were not part of the training. For the synthetic rubber material, ground-truth stresses can be computed for arbitrary deformations, which allows evaluation on Treloar's invariant plane \cite{treloar2005physics}. The first and second invariants span the x- and y-axes; the plane ranges from uniaxial to equibiaxial tension, with pure shear at the angular midpoint (note that the two axes use different scales). Only these three highlighted loading paths are used during training, so every point in between or beyond corresponds to a loading state that the model has never seen. This setup lets us probe both generalization (to intermediate loading states) and extrapolation (beyond the training range).

Figure \ref{fig_predictions_curves_baseline} d) shows that the baseline model generalizes well to unseen loading conditions but has small difficulties when extrapolating significantly beyond the training data. The relative errors remain well under 5\%, but they are clearly noticeable. The models generated by our pipeline, in contrast, generalize and extrapolate remarkably well: within the range evaluated here, their accuracy does not drop noticeably when moving away from the training loading paths or beyond the training range.

\subsection{Are the generated models comparable in parameter counts to the baselines?}\label{sec_Results_network_size}

\begin{table}[h]
\caption{Architectural specifications of constitutive artificial neural networks (CANNs) used as baseline and generated by Creator–Inspector (Claude Opus 4.7 and Kimi K2.5, no tool access), with predictions shown in Figures~\ref{fig_predictions_curves_opus} - \ref{fig_predictions_curves_baseline}.}
\label{tab_hidden_layers}
\centering
\renewcommand{\arraystretch}{1.3} % extra row height
\begin{tabular}{|>{\raggedright\arraybackslash}p{0.28\linewidth}
                >{\raggedright\arraybackslash}p{0.27\linewidth}
                >{\centering\arraybackslash}p{0.35\linewidth}|}
\hline
\textbf{Material} & \textbf{Model} & \textbf{Neurons per hidden layer} \\
\hline
Brain tissue  & Opus-generated & 12, 12 \\
              & Kimi-generated & 128, 128, 128, 128 \\
              & Baseline       & 100 \\
Rubber        & Opus-generated & 20, 20 \\
              & Kimi-generated & 10, 10 \\
              & Baseline       & 16, 16 \\
Synth. rubber & Opus-generated & 24, 12 \\
              & Kimi-generated & 256, 256 \\
              & Baseline       & 16, 16 \\
Skin          & Opus-generated & 48 \\
              & Kimi-generated & 112, 112 \\
              & Baseline       & 8, 16 \\
\hline
\end{tabular}
\end{table}

A recurring concern when comparing the accuracy of different constitutive models is that parameter counts should be matched. It is worth emphasizing that the goal of this work is not to match or beat the accuracy of human-expert-designed models, but rather to understand what a multi-agent system powered by today's frontier LLMs can do in terms of producing both accurate and physically meaningful constitutive models. Still, a brief look at network sizes helps put the reported accuracies into perspective.

In the GenCANN work \cite{tacke2025automating}, the LLM was, in one variant of the setup, explicitly instructed to reproduce the baseline's network size so as to enable a head-to-head comparison. In the realistic setting targeted here — where either a single LLM or, as in our case, a specialized multi-agent system designs a constitutive model for a new material for which no established model exists — such a head-to-head constraint is not meaningful. The useful question is rather whether restricting the network size damages the LLM's ability to design a working model. The answer given in \cite{tacke2025automating} is clear: the accuracy loss from restricting the network size is small, and the LLMs are robust in this regard. Since this question has already been settled, we did not repeat the corresponding experiments here.

For transparency, however, we report the network sizes of the models whose predictions are shown in Figures~\ref{fig_predictions_curves_opus}--\ref{fig_predictions_curves_baseline} in Table~\ref{tab_hidden_layers}. Two observations stand out. First, the Creator-Inspector pair powered by Claude Opus, entirely by its own design choices, tends to produce networks of a size comparable to the baselines. Small networks are always welcome in terms of runtime, so this is a favorable characteristic. Second, the Kimi-powered pair tends to produce larger networks than the baseline. This is consistent with the observations in \cite{tacke2025automating}, where unconstrained LLMs showed a tendency toward oversized architectures.

The decisive criterion, however, is not the absolute network size but whether the generated models overfit the small training datasets. As the generalization and extrapolation results in Section~\ref{sec_Results_extrapolation} demonstrate, they clearly do not. The network sizes chosen by the agents are therefore appropriate for the task, and the experiments in \citep{tacke2025automating} make us confident that comparable accuracy could be reached with smaller networks if desired.

\subsection{How reliably do the generated models reach the presented accuracy?}\label{sec_Results_variance}

\begin{figure}[h]
  \centering
  \includegraphics[width=\linewidth]{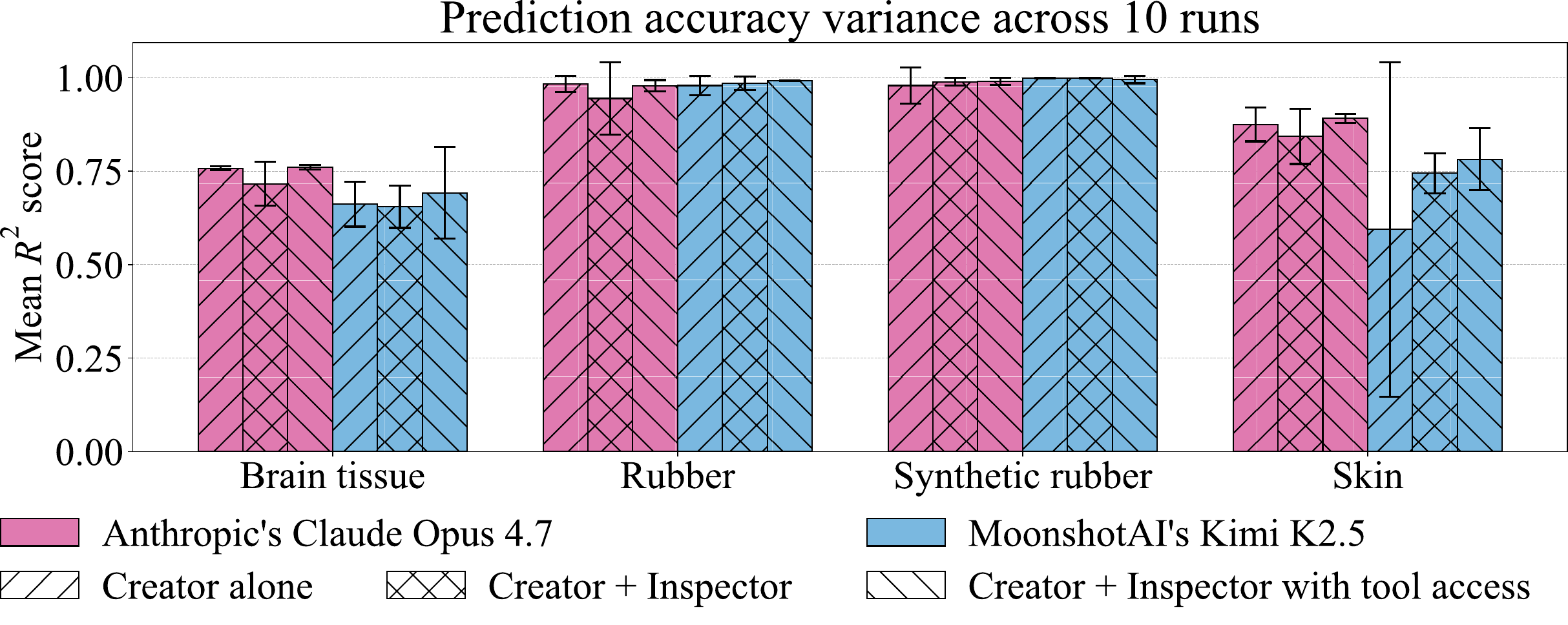}
  \caption{Reliability of the achieved accuracy across runs. Mean $R^2$ score of model performance, averaged over the loading conditions of each dataset and over ten independent runs and showing also standard deviation.}
  \label{fig_predictions_accuracy_variance}
\end{figure}

Since the predictions shown so far correspond to the best out of ten runs, a natural follow-up question is how reliably the Creator–Inspector pair produces models of comparable accuracy. Figure \ref{fig_predictions_accuracy_variance} answers this. Each bar shows the mean $R^2$ score, averaged over the loading conditions of the respective dataset and over ten runs, with the standard deviation indicated by a line on top. We report this for every configuration: Creator alone, Creator with Inspector, and Creator with tool-enabled Inspector, each for Claude Opus 4.7 and Kimi K2.5, on every dataset.

On both rubber datasets, the variance between runs is close to zero: virtually every run reaches near-perfect accuracy, so the pipeline very reliably delivers highly accurate models. The one exception is the Opus-powered Creator–Inspector pair on the experimental rubber dataset, where a single outlier run with an $R^2$ score of 0.66 slightly lowers the mean and inflates the standard deviation. Since the pipeline is partly stochastic, such individual cases can occur. On the brain tissue data, the variance is moderately larger but still modest, and it is higher for the Kimi-powered agents than for the Opus-powered ones. The skin data show variance at a similar level, with Opus being both more accurate and more reliable than Kimi throughout. The striking exception is the Kimi-powered Creator alone, whose mean drops to 0.59 with a standard deviation of 0.47, driven by individual runs that only reach poor accuracy. For Kimi on skin, the accuracy increases when the Inspector is added and increases again when the Inspector is granted tool access. Since Kimi performs overall more poorly than Opus and skin is certainly the most challenging material to fit in this study, we may hypothesize that when the Creator is particularly challenged by the modeling task, adding the Inspector helps not only with constraint adherence but even with accuracy: at this frontline, the second agent acts not as a trade-off but as a complement. Across all datasets and both backbones, we conclude that the accuracies shown in Figures \ref{fig_predictions_curves_opus} - \ref{fig_predictions_curves_baseline} are not the results of lucky individual runs but are representative of what the pipeline delivers in a typical execution.

\subsection{How effective is the iterative refinement of the models?}\label{sec_Results_refinement}

\begin{figure}[t]
  \centering
  \includegraphics[width=\linewidth]{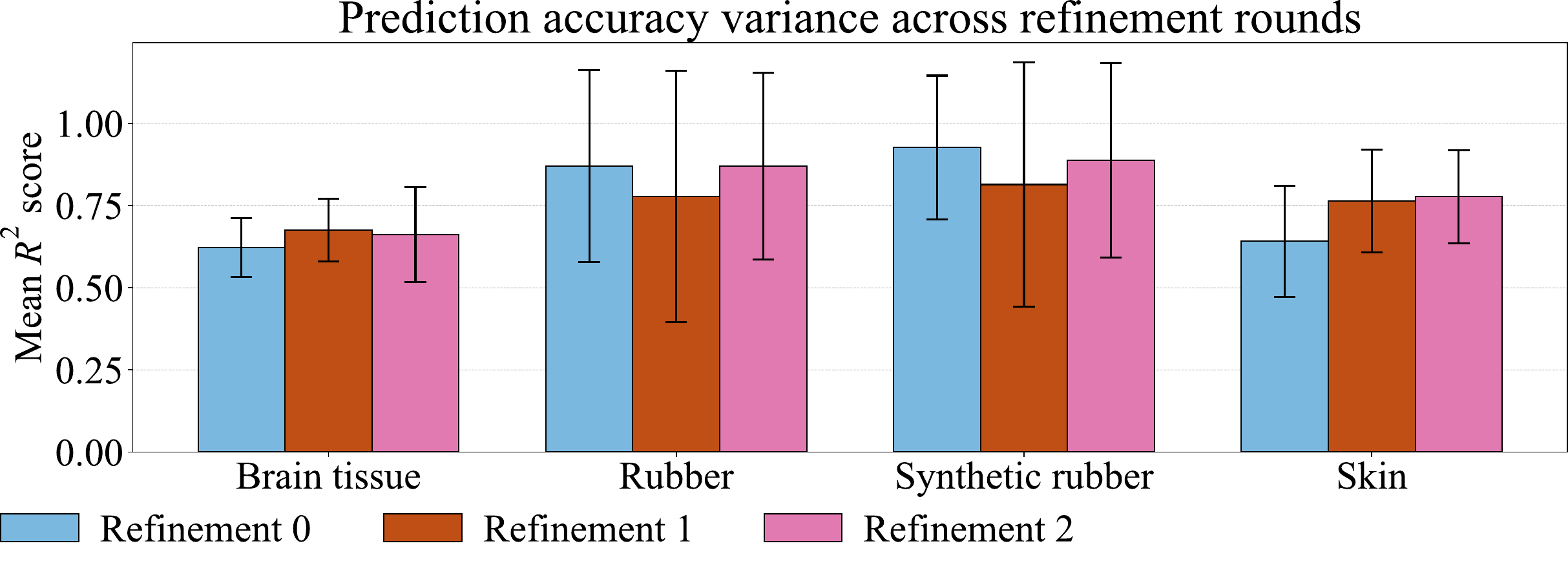}
  \caption{Effect of the iterative refinement loop. Mean $R^2$ score (including standard deviation) of the generated model per refinement round, pooled across all six configurations (Creator alone, Creator with Inspector, and Creator with tool-enabled Inspector, each powered by Claude Opus 4.7 and Kimi K2.5) and shown separately for each dataset. Negative average $R^2$ scores were clipped to zero to keep extreme outliers from dominating the plot.}
  \label{fig_predictions_refinement_variance}
\end{figure}

Since our pipeline includes a refinement loop, we should check whether that loop actually has an effect on the generated models. To this end, we pool the six configurations analyzed previously — Creator alone, Creator with Inspector, and Creator with tool-enabled Inspector, each for Opus and Kimi — into a single group of experiments, but separate the results by refinement round and by dataset. The corresponding accuracies are shown in Figure \ref{fig_predictions_refinement_variance}. For readability, we clip negative average $R^2$ scores to zero, as a small number of extreme outliers would otherwise dominate the plot. This clipping was not necessary in Figure \ref{fig_predictions_accuracy_variance}, because the refinement process itself filtered out these drastic errors — already a first indication that the loop gives the LLMs room to experiment and to discard clearly broken models before the final export.

The variance across runs in Figure \ref{fig_predictions_refinement_variance} is visibly larger than in Figure \ref{fig_predictions_accuracy_variance}. For the two rubber datasets, no clear trend across the three refinement rounds is discernible: the standard deviations are simply too large to support any significant statement. For the brain tissue data, the variance is much smaller, and a small improvement after the first refinement is visible, while the second refinement does not yield any further improvement. For the skin data, the picture is clearest: the mean accuracy rises with each refinement round, while the variance even decreases slightly.

Taken together, these patterns suggest that the agents use the refinement rounds for both exploration and exploitation, depending on how successful the initial models already are. On the rubber datasets, where the agents reliably produce at least one highly accurate model early on — as reflected in the tiny variance of the final results in Figure \ref{fig_predictions_accuracy_variance} — the remaining rounds are evidently used to try out more unconventional ideas, which inflates the per-round variance. On the brain tissue data, where obtaining an accurate model is harder, the agents also explore but stay more focused: the variance is larger than in Figure \ref{fig_predictions_accuracy_variance} but visibly smaller than for rubber in Figure \ref{fig_predictions_refinement_variance}, and the small improvement after the first refinement shows that incremental progress does occur where it is strongly needed. The skin data confirm this pattern most convincingly: on the most challenging dataset of this study, the refinement rounds deliver the largest incremental gains, raising the mean accuracy round by round. Refinement thus pays off most where the initial proposals leave the most room for improvement. In all cases the refinement loop is used productively, just in different ways depending on how quickly the Creator arrives at highly accurate models.

\subsection{What does the Inspector cost?}\label{sec_Results_cost}

Adding a second agent to the pipeline naturally increases its cost compared to the Creator alone: every inspection consumes additional tokens, and every violation flag triggers an additional creation round and thus an additional CANN training. Since CANN trainings complete within seconds to a few minutes on consumer hardware, we simply count them. The frontier LLMs, in contrast, run on external servers, so we count the tokens they consume and the cost these translate to.

For Claude Opus 4.7, the Inspector without tool access adds roughly 17,000 input and 6,500 output tokens per run (+58\% in total compared to the Creator alone), corresponding to 0.25 USD at current provider prices, and it triggers roughly one additional CANN training per run. With tool access, the addition amounts to roughly 28,000 input and 4,000 output tokens (+79\%) at a similar extra cost of 0.24 USD, while additional trainings become rare at 0.08 per run on average. For Kimi K2.5, the Inspector without tools adds roughly 12,000 input and 4,500 output tokens (+52\%), corresponding to about one cent, and 0.2 additional trainings per run on average; with tool access, the extra volume grows to roughly 63,000 input and 11,000 output tokens (+229\%), corresponding to five cents and, on average, 1.6 additional trainings per run.

These additional costs are moderate in relative terms and negligible in absolute ones: even the most expensive configuration remains below one USD per complete run — orders of magnitude below the cost of the expert time it saves.

\section{Discussion and Conclusions} \label{sec_Discussion}

\subsection{Effectiveness of the multi-agent setup:\\two agents are better than one}

The central methodological contribution of this work is the Creator–Inspector paradigm. Our analysis of the state transitions between Creator and Inspector shows that the two agents communicate effectively. For the Opus-powered pair, whenever the Inspector raises a new violation, the Creator resolves it in the very next iteration in 90\% of cases, and violations persist in only 5\%. For the Kimi-powered pair, still 72\% of new violations are resolved directly and only 17\% persist. In other words, when the Inspector sees the need to intervene, its feedback is actionable rather than merely critical. This is the behavior one would hope for from a generator–critic pair, and it mirrors the empirical case that has been made for multi-agent LLM systems more broadly, where a clean separation between proposing and inspecting roles consistently outperforms monolithic chain-of-thought pipelines \citep{du2024improving, gridach2025agentic, chen2024autoagents, liu2024a}. For constitutive modeling, however, no prior single-agent system — SGA \citep{ma2024llm}, CSGA \citep{tacke2025constitutive}, or GenCANN \citep{tacke2025automating} — had closed this gap.

The effect of this collaboration on the final output is striking. Introducing the Inspector raises the share of proposed models that pass the numerical validation of all nine physical constraints from 90\% to 95\% for the Creator powered by Claude Opus 4.7, and from 47\% to 60\% for the Creator powered by Kimi K2.5. The characteristics of any specific LLM are of limited interest to us, as frontier models evolve too quickly to make claims tied to a particular release. What matters is that the same effect emerges in two LLMs that otherwise behave very differently, so the improvement can be attributed to the paradigm itself rather than to any model-specific behavior. At the same time, agentic pipelines driven by LLMs remain inherently stochastic at inference, which makes hard guarantees difficult to achieve. Our sampling-based validators provide only strong empirical evidence rather than mathematical proof of admissibility, and even the best configuration approaches perfect adherence without fully reaching it. Already at its current level, however, the approach serves the purpose it was designed for: motivated by accessibility, it offers users who have neither an established model for their material at hand nor the expertise to design one a physically validated model that would otherwise be out of reach. In Subsections \ref{subsec_discussion_autonomy} and \ref{subsec_discussion_usability}, we present two arguments why we expect the reliability of the process to increase even further in the future. 

Reliability is only one side of the coin; accuracy is the other. Figures \ref{fig_predictions_curves_opus}–\ref{fig_predictions_accuracy_variance} show that representative models produced by our pipeline not only withstand all numerical constraint checks but are also highly accurate, with the Opus-powered models reaching performance nearly on par with or even surpassing the models designed by human experts. Equally important, both the Opus- and Kimi-powered models generalize and extrapolate remarkably well beyond the training loading conditions, as evidenced by their predictions on Treloar's invariant plane. Together, this combination of physical validity, high accuracy, and strong extrapolation makes the generated models directly usable across a wide range of applications.

Beyond these concrete results, it is worth stressing that the paradigm itself is deliberately technique-agnostic. Our use of CANNs is a concrete demonstration vehicle, chosen because CANNs are well established and offer a favorable trade-off between flexibility and physical structure, but the same pipeline could be instantiated around PANNs, CKANs, or even purely symbolic laws without conceptual changes — only the prompts given to the Creator and Inspector would need to be adapted. In this sense, the Creator–Inspector paradigm closes the gap left open by GenCANN — which produced accurate but physically unchecked models — and provides a general template for physics-aware LLM-driven constitutive modeling.

\subsection{Autonomy of the agents:\\calibrated to match what today's LLMs can reliably deliver}\label{subsec_discussion_autonomy}

Agentic LLM-powered systems can be more or less autonomous, and it is worth being explicit about where our pipeline sits on this spectrum. Two design choices constrain the autonomy of the agents: the alternation between Creator and Inspector is predefined rather than negotiated by the agents themselves, and the numerical validation tools available to the Inspector are hard-coded rather than implemented on the fly. Within these constraints, however, both agents retain meaningful freedom. The Creator autonomously decides on the architecture, the invariant-based building blocks, the activation functions, the handling of the fiber direction in the transversely isotropic case, and the training procedure of each proposed CANN. The Inspector decides how many creation updates are triggered - implicitly, through its verdicts - and, in the tool-enabled variant, which tools to invoke and how many. On balance, the Creator is best described as constrained but expressive, and the Inspector as semi-autonomous.

Autonomy, however, is not a value in itself; it is only useful insofar as it translates into a practical advantage. The two constraints we imposed are motivated precisely by this consideration. Fixing the Creator-Inspector alternation reflects the fact that, at the current level of LLM reliability, it is simply not safe to let the Creator decide whether inspection is needed: our results in Section \ref{sec_Results_verification} show that even a frontier model such as Claude Opus 4.7 produces physically invalid models in roughly one out of ten proposals when left unchecked. If, at some point in the future, Creators become reliable enough to estimate their own constraint adherence, one could naturally relax this constraint and let the Creator call the Inspector on demand. A similar argument applies to the hard-coded tools: letting the Inspector implement numerical validation routines on the fly would currently add a new and uncontrolled source of errors, but once LLMs routinely produce correct and trustworthy validation code, the external tools would largely lose their purpose anyway. The balance between autonomy and reliability that we chose here is therefore not meant as a universal prescription, but as the one that, in our view, best matches the current capabilities of the underlying LLMs.

\subsection{Practical usability:\\scales automatically with LLM and agentic progress}\label{subsec_discussion_usability}

What does this mean for someone who wants to use the pipeline in practice? Our experiments can be read, in part, as a capability study: they probe how far today's frontier LLMs can be pushed, within a carefully designed agentic setup, toward the task of generating physically admissible constitutive models. The answer is encouraging. For hyperelastic, incompressible materials, both isotropic and transversely isotropic, the Creator–Inspector pair powered by Claude Opus 4.7 is ready to produce constitutive models that respect the relevant physical constraints, are highly accurate, and generalize to unseen loading conditions as well as extrapolate beyond the training data. In the future, the same approach can be tested on a broader range of material classes, such as viscoelastic materials. A further practical strength is the form in which the models are returned: plain Python files that can be inspected, modified, and reused directly. And should the user wish to modify the model or any aspect of it, a plain-text prompt is enough to request the change.

Two developments are likely to push this horizon further still. The first concerns the agentic design itself. Our pipeline is, to the best of our knowledge, the first multi-agent system for constitutive modeling, and there is substantial room for improvement through additional agents, longer refinement loops, and more carefully engineered prompts. The impact of such refinements should not be underestimated: on the widely used HumanEval coding benchmark, GPT-3.5 reaches 48.1\% in a zero-shot setting and GPT-4 reaches 67.0\%, but wrapping GPT-3.5 in an iterative agentic workflow pushes it to 95.1\% \citep{Ng2024AgenticDesignPatterns} — an improvement that dwarfs the underlying model upgrade. In other words, the agentic scaffolding around an LLM can matter more than the LLM itself, and the design space we have begun to explore here is still largely open.

The second development concerns the LLM backbones. As they continue to improve, we expect both the accuracy of the generated models and their adherence to physical constraints to rise with them, and we expect the Creator and Inspector to become trustworthy enough to operate with increasing autonomy. Our results place Claude Opus 4.7 clearly ahead of Kimi K2.5 today, but given how rapidly open-weight capability is advancing, we expect open-weight models to soon reach the level at which Opus 4.7 operates now. Switching to a more capable backbone is a one-line change in our codebase, so the pipeline is well positioned to ride this curve rather than fight it. In the spirit of Sutton's bitter lesson — that general methods which scale with computation eventually outperform approaches relying on hand-engineered domain knowledge \citep{Sutton2019BitterLesson} — the Creator–Inspector paradigm is deliberately built to benefit from stronger LLMs rather than to depend on any particular one.

\newpage

\section*{Declarations} \label{sec_declarations}

\subsection*{Software and data availability} \label{sec_software_data_availability}

The human brain tissue and porcine skin tissue datasets and the corresponding baseline CANN implementations are available at \href{https://github.com/LivingMatterLab/CANN}{https://github.com/LivingMatterLab/CANN}. The rubber datasets and their corresponding baseline CANN implementation can be found at  \href{https://github.com/ConstitutiveANN/CANN}{https://github.com/ConstitutiveANN/CANN}. The code for the approach presented in this work, together with all artifacts reported here, including the full set of prompts and the complete conversation logs between the Creator and Inspector agents, is available at \href{https://github.com/AgenticModeling/MultiAgentConstitutiveModeling26}{https://github.com/AgenticModeling/MultiAgentConstitutiveModeling26}.

\subsection*{Author contributions} \label{sec_author_contribution}

Marius Tacke conceived and designed the study, developed the methodology and software, curated the data, and carried out the investigation, validation, and formal analysis. He also wrote the original draft. Matthias Busch contributed to the formal analysis and prepared visualizations. Kian Abdolazizi contributed to the methodology and validation. Jonas Eichinger contributed to the methodology. Kevin Linka contributed to the conceptualization of the study. Roland Aydin contributed to the conceptualization and visualization and supervised the work. Christian Cyron provided resources, contributed to the visualization, supervised the work, and administered the project. All authors reviewed and edited the manuscript.

\subsection*{Funding} \label{sec_funding}

Funded by the Deutsche Forschungsgemeinschaft (DFG, German Research Foundation) under Germany's Excellence Strategy – EXC 3120/1 – 533771286. Christian Cyron gratefully acknowledges support of the European Research Council (ERC) under the European Union’s Horizon Europe research and innovation programme (grant agreement No 101167207 / MechVivo).

\subsection*{Competing interests} \label{sec_competing_interests}
The authors declare no competing interests.

\subsection*{Ethical approval} \label{sec_ethics}

No animals or human subjects were used in this study. All data are taken from the literature.

\newpage

\bibliography{references}
\bibliographystyle{references}

\newpage

\appendix

\section{Numerical validation of the physical constraints}\label{sec_Appendix_validations}

This appendix specifies the numerical validators described in Section~\ref{sec_Background_constraints} in sufficient detail for an independent reimplementation. All validators receive a trained model and return a binary verdict (passed or failed). A constraint is reported as passed only when no violation is found across all sampled states. Most validators perform element-wise comparisons through a common helper that combines a relative tolerance of $\tau_\mathrm{rel}=10^{-3}$ — i.e., agreement to three significant digits of the larger argument — with an absolute floor of $\tau_\mathrm{abs}=10^{-4}$ that takes over for values near zero, where any relative tolerance would collapse to floating-point noise. A few simpler validators (energy and stress normalization, non-negativity of the strain energy) instead apply a plain absolute tolerance of $\tau = 10^{-3}$. These choices are a deliberate compromise between sensitivity to genuine violations and robustness to numerical noise. Three validators — material symmetry, ellipticity, and thermodynamic consistency — additionally depend on the material's symmetry class, and the base sampling set of deformation gradients is augmented in the transversely isotropic (TI) case; we give both the isotropic and the transversely isotropic specifications below.

\subsection{Common sampling spaces}\label{app_sampling}

\paragraph{Plane-strain deformation gradients.}
Because the materials considered here are incompressible, all admissible deformation gradients satisfy $\det\mathbf{F}=1$. We work with plane-strain deformations of the form
\begin{equation}
\mathbf{F} = \begin{pmatrix} F_{11} & F_{12} & 0 \\ F_{21} & F_{22} & 0 \\ 0 & 0 & 1/(F_{11}F_{22}-F_{12}F_{21}) \end{pmatrix},
\end{equation}
which guarantees $\det\mathbf{F}=1$ and matches the experimental loading scenarios (uniaxial, biaxial, pure and simple shear). The base sampling set $\mathcal{F}_\mathrm{base}$ contains the identity, uniaxial tension and compression states with stretches $\lambda\in\{0.5,0.7,0.9,1.1,1.3,1.5,2.0,3.0\}$, equibiaxial states with $\lambda\in\{0.7,0.9,1.1,1.3,1.5,2.0\}$, planar (pure) shear states with $\lambda\in\{0.5,0.7,0.9,1.1,1.5,2.0,3.0\}$, simple shear states with $\gamma\in\{-1.0,-0.5,-0.1,0.1,0.5,1.0\}$, off-equibiaxial combinations, and a few combined stretch--shear states. In total, $|\mathcal{F}_\mathrm{base}|\approx 35$ representative states. For the transversely isotropic problem, plane-strain states alone leave $C_{13}=C_{23}=0$, so an in-plane fiber would never be exposed to the out-of-plane shear modes to which the pseudo-invariant $I_5=\mathrm{cof}(\mathbf{C}):\mathbf{N}$ couples. The base set is therefore augmented by the out-of-plane simple-shear states $\mathbf{F}=\mathbf{I}+\gamma\,\mathbf{e}_i\otimes\mathbf{e}_j$ with $\gamma\in\{\pm 0.3,\pm 0.7\}$ and $(i,j)\in\{(1,3),(2,3),(3,1),(3,2)\}$, together with two combined stretch--shear states $\mathbf{F}=\mathrm{diag}(\lambda,1,\lambda^{-1})+0.5\,\mathbf{e}_1\otimes\mathbf{e}_3$ with $\lambda\in\{1.5,0.7\}$, all satisfying $\det\mathbf{F}=1$. Every validator that samples $\mathcal{F}_\mathrm{base}$ then uses this augmented set $\mathcal{F}_\mathrm{base}^\mathrm{TI}$ with $|\mathcal{F}_\mathrm{base}^\mathrm{TI}|\approx 53$ states.

\paragraph{Three-dimensional rotations.}
Out-of-plane rotations are sampled as $\mathbf{Q}=\mathbf{R}_x(\varphi_1)\mathbf{R}_y(\varphi_2)\mathbf{R}_z(\varphi_3)$, where $\mathbf{R}_x,\mathbf{R}_y,\mathbf{R}_z$ denote the elementary rotations about the coordinate axes. We use a fixed set $\Phi$ of nine angle triples chosen to combine simple multiples of $\pi/2,\pi/3,\pi/4,\pi/6$ with angles that are not rational multiples of $\pi$ for generality, yielding the proper-rotation set $\mathcal{Q}^+$. Reflections are obtained by post-multiplying with $\mathrm{diag}(1,1,-1)$, yielding the set $\mathcal{Q}^- = \{\mathbf{Q}\,\mathrm{diag}(1,1,-1)\,:\,\mathbf{Q}\in\mathcal{Q}^+\}$ for the isotropic material-symmetry test. In-plane rotations $\mathbf{Q}_\mathrm{2D}(\theta)$ about the $z$-axis are reused as waypoint constructions in the thermodynamic-consistency loops, for the fiber recovery, and for the orientation sweep in the transversely isotropic ellipticity sampling.

\subsection{Thermodynamic consistency}\label{app_thermo}

Following the path-independence approach of \citet{upadhyay2019thermodynamics, gonzalez2019thermodynamically}, we test thermodynamic consistency by evaluating four metrics: two on closed deformation loops, and two on a pair of open paths that share the same end states.

\paragraph{Paths.}
Closed loops and one two-path comparison are constructed by piecewise-linear interpolation between waypoints in $(F_{11},F_{12},F_{21},F_{22})$ space -- extended by $(F_{13},F_{23})$ for the transversely isotropic problem -- with the out-of-plane stretch recomputed at each step to enforce $\det\mathbf{F}=1$. Each segment is discretized into $n_\mathrm{seg}=200$ steps.
\begin{itemize}
    \item Loop~1 uses purely diagonal waypoints and traverses uniaxial tension, planar tension, and equibiaxial states before returning to the identity.
    \item Loop~2 additionally activates off-diagonal entries through simple shear and includes a waypoint obtained by applying the in-plane rotation $\mathbf{Q}_\mathrm{2D}(\pi/4)$ to a stretch--shear state, exercising the full chain rule through the invariants.
    \item Loop~3, used only for the transversely isotropic problem, additionally ramps the out-of-plane shear components $F_{13}$ and $F_{23}$, combined with in-plane stretch and shear. Since $F_{31}=F_{32}=0$, these components do not enter $\det\mathbf{F}$, and the construction remains volume-preserving. This loop closes a blind spot of the plane-strain loops, which keep $\mathrm{d}F_{13}=\mathrm{d}F_{23}=0$ and are therefore insensitive to non-gradient stress contributions confined to the out-of-plane components.
    \item Path~A connects $\mathbf{I}$ to a uniaxial tension state directly. Path~B connects the same end points by first loading through pure shear and then adjusting the transverse stretch at constant axial stretch.
\end{itemize}

\paragraph{Metrics.}
At each discretization step $k$, the model returns the energy $\Psi_k$ and the stress $\mathbf{P}_k$. The incremental and cumulative work are computed via the trapezoidal rule,
\begin{equation}
\delta W_k \;=\; \tfrac{1}{2}\bigl(\mathbf{P}_k+\mathbf{P}_{k+1}\bigr):\Delta\mathbf{F}_k,
\qquad W_k = \sum_{j=0}^{k-1}\delta W_j.
\end{equation}
On each \emph{closed loop}, two checks are performed:
\begin{enumerate}
    \item \emph{Normalized loop residual.} The net work over the loop must 
    vanish relative to the total work expended,
    \begin{equation}
    \eta \;=\; |W_N|\Big/\sum_k |\delta W_k| \;\leq\; \tau_\mathrm{loop},
    \qquad \tau_\mathrm{loop}=10^{-2}.
    \end{equation}
    The tolerance here is looser than the one used elsewhere because trapezoidal integration accumulates discretization error along the full path length, so a stricter bound would conflate genuine thermodynamic violations with quadrature noise.
    \item \emph{Stress uniqueness at loop closure.} The model must return to 
    the same stress when the deformation returns to its starting point,
    \begin{equation}
    \mathbf{P}(\mathbf{F}_\mathrm{start}) \;\approx\; \mathbf{P}(\mathbf{F}_\mathrm{end}),
    \end{equation}
    componentwise to relative tolerance $\tau_\mathrm{rel}$ with absolute floor $\tau_\mathrm{abs}$. Being a direct model evaluation rather than an integrated quantity, this check is independent of any quadrature drift and specifically catches hidden statefulness in the implementation.
\end{enumerate}
On the \emph{two-path comparison}, two further checks are performed:
\begin{enumerate}\setcounter{enumi}{2}
    \item \emph{Two-path agreement at the common end state.} The cumulative 
    work along Path~A and Path~B must coincide at the shared end state, 
    $W_N^A \approx W_N^B$, to relative tolerance $\tau_\mathrm{rel}$ with 
    absolute floor $\tau_\mathrm{abs}$.
    \item \emph{Pointwise work--energy consistency.} Along each open path, 
    the cumulative work must reproduce the energy difference at every step,
    \begin{equation}
    \bigl|W_k - (\Psi_k-\Psi_0)\bigr| \;\leq\; \tau_\mathrm{rel}\,
    \max\bigl(|W_k|,\,|\Psi_k-\Psi_0|\bigr),
    \qquad \tau_\mathrm{rel}=10^{-2}.
    \end{equation}
\end{enumerate}

The validator returns ``passed'' only if all four metrics are satisfied
for every loop and for the two-path pair.

\subsection{Stress symmetry}\label{app_stress_sym}

For every $\mathbf{F}\in\mathcal{F}_\mathrm{base}$ the model is queried 
for the predicted stress $\mathbf{P}$, the Cauchy stress is computed as 
$\boldsymbol{\sigma}=\mathbf{P}\mathbf{F}^T$ (using $J=1$), and the 
condition $\boldsymbol{\sigma}=\boldsymbol{\sigma}^T$ is verified 
componentwise to relative tolerance $\tau_\mathrm{rel}$ with absolute 
floor $\tau_\mathrm{abs}$ for every sampled $\mathbf{F}$.

\subsection{Objectivity}\label{app_objectivity}

We verify
\begin{equation}
\Psi(\mathbf{Q}\mathbf{F}) \;=\; \Psi(\mathbf{F})
\qquad \forall\,\mathbf{F}\in\mathcal{F}_\mathrm{base},\ \forall\,\mathbf{Q}\in\mathcal{Q}^+,
\end{equation}
to relative tolerance $\tau_\mathrm{rel}$ with absolute floor 
$\tau_\mathrm{abs}$. Since $\Psi$ is computed independently of the 
Lagrange-multiplier pressure $p$, the plane-stress assumption 
$P_{33}=0$ does not enter this check, and arbitrary three-dimensional 
rotations $\mathbf{Q}\in\mathcal{Q}^+$ can be used. This holds for any 
material symmetry class, as $\mathbf{C}=\mathbf{F}^T\mathbf{F}$ (and hence 
every invariant of $\Psi$) is unchanged by a rotation from the left.

\subsection{Material symmetry}\label{app_mat_sym}

\paragraph{Isotropic materials.}
The symmetry group is $\mathcal{O}(3)$. The validator checks
\begin{equation}
\Psi(\mathbf{F}\mathbf{Q}^T) \;=\; \Psi(\mathbf{F})
\qquad \forall\,\mathbf{F}\in\mathcal{F}_\mathrm{base},\ \forall\,\mathbf{Q}\in\mathcal{Q}^+\cup\mathcal{Q}^-,
\end{equation}
to relative tolerance $\tau_\mathrm{rel}$ with absolute floor 
$\tau_\mathrm{abs}$, again using the fact that $\Psi$ does not depend on $p$.

\paragraph{Transversely isotropic materials.}
The symmetry group reduces to the transverse-isotropy group about the fiber direction $\mathbf{n}$, i.e. the set $\{\mathbf{Q}\in\mathcal{O}(3):\mathbf{Q}\mathbf{n}=\pm\mathbf{n}\}$. Every such $\mathbf{Q}$ satisfies $\mathbf{Q}^T\mathbf{N}\mathbf{Q}=\mathbf{N}$ with the structure tensor $\mathbf{N}=\mathbf{n}\otimes\mathbf{n}$, and therefore leaves the fiber pseudo-invariants $I_4=\mathbf{C}:\mathbf{N}$ and $I_5=\mathrm{cof}(\mathbf{C}):\mathbf{N}$ -- and hence $\Psi$ -- unchanged, where for $I_5$ we additionally use $\mathrm{cof}(\mathbf{Q}\mathbf{C}\mathbf{Q}^T)=\mathbf{Q}\,\mathrm{cof}(\mathbf{C})\,\mathbf{Q}^T$ for orthogonal $\mathbf{Q}$. Because $\mathbf{n}$ is learned by the model rather than prescribed, the validator proceeds in two steps.

\emph{Fiber recovery.} The fiber is assumed to lie in the $xy$-plane. We evaluate the energy on the planar probes
\begin{equation}
\mathbf{U}(\varphi)=\mathbf{Q}_\mathrm{2D}(\varphi)\,\mathbf{D}\,\mathbf{Q}_\mathrm{2D}(\varphi)^T,
\qquad \mathbf{D}\in\bigl\{\mathrm{diag}(\lambda,\lambda^{-1},1),\ \mathrm{diag}(\lambda,1,\lambda^{-1})\bigr\},
\qquad \lambda\in\{1.3,\,2.0,\,3.0\},
\end{equation}
at $n_\varphi=36$ equispaced angles $\varphi\in[0,\pi)$, and read candidate axis angles from the Fourier modes
\begin{equation}
\theta=\tfrac{1}{p}\arg\!\Big(\textstyle\sum_k \Psi(\mathbf{U}(\varphi_k))\,e^{\,p\,i\varphi_k}\Big),
\end{equation}
with $p=2$ for both probe families and additionally $p=4$ for the reciprocal family $\mathrm{diag}(\lambda,\lambda^{-1},1)$; per signal, the phase is taken from the amplitude $\lambda$ with the largest mode magnitude, so that fiber contributions that only activate at larger stretches are detected as well. Combining two probe families and two modes is necessary because each signal alone has a blind spot. On the reciprocal family, $\mathrm{cof}(\mathbf{C})\!:\!\mathbf{N}$ coincides with $\mathbf{C}\!:\!\mathbf{M}$ formed with the in-plane normal $\mathbf{m}$, so a model with balanced $I_4$- and $I_5$-channels is $\pi/2$-periodic over $\varphi$ and its second mode vanishes identically; the fourth mode still carries the axis, with a provably non-negative coefficient for the convex, non-decreasing channels mandated here, so that its phase is unambiguous up to $\pi/2$. Balanced channels that are furthermore linear render $\Psi$ constant on the reciprocal family altogether, which the non-reciprocal family $\mathrm{diag}(\lambda,1,\lambda^{-1})$ resolves. Since each recovered $\theta$ determines the axis only up to its in-plane normal, both $(\cos\theta,\sin\theta,0)$ and $(\cos(\theta+\tfrac{\pi}{2}),\sin(\theta+\tfrac{\pi}{2}),0)$ are retained, yielding up to six candidate axes after deduplication.

\emph{Symmetry set and verdict.} About a candidate axis $\mathbf{n}$ (with in-plane normal $\mathbf{m}$) we assemble
\begin{equation}
\mathcal{Q}^\mathrm{TI}(\mathbf{n}) \;=\; \{\mathbf{R}_{\mathbf{n}}(\alpha):\alpha\in\mathcal{A}\}\;\cup\;\{\mathbf{I}-2\,\mathbf{n}\otimes\mathbf{n},\ \mathbf{I}-2\,\mathbf{m}\otimes\mathbf{m}\},
\end{equation}
where $\mathbf{R}_{\mathbf{n}}(\alpha)$ is the rotation about $\mathbf{n}$ by $\alpha$, $\mathcal{A}=\{\pi/6,\pi/4,\pi/3,\pi/2,2\pi/3,\pi,0.73\}$ combines simple rational multiples of $\pi$ with an angle that is not a rational multiple of $\pi$ for generality (analogously to the three-dimensional rotation set), and the two reflections map the fiber to $-\mathbf{n}$ and $+\mathbf{n}$, respectively. The validator checks $\Psi(\mathbf{F}\mathbf{Q}^T)=\Psi(\mathbf{F})$ (to $\tau_\mathrm{rel}$, $\tau_\mathrm{abs}$) for all $\mathbf{F}\in\mathcal{F}_\mathrm{base}^\mathrm{TI}$ and $\mathbf{Q}\in\mathcal{Q}^\mathrm{TI}(\mathbf{n}_i)$, and returns ``passed'' if invariance holds for at least one candidate axis $\mathbf{n}_i$. Note that the check verifies invariance under the transverse-isotropy group, i.e., that the model possesses at least this symmetry: the true fiber axis passes, and a model of higher symmetry (e.g. a fully isotropic one, which is invariant about any axis) passes correctly as well, whereas a model of lower symmetry (e.g. orthotropic), which is invariant under no continuous rotation about either candidate, is correctly rejected.

\subsection{Ellipticity}\label{app_ellipticity}

We verify the rank-one convexity (Legendre--Hadamard) condition
\begin{equation}\label{eq_app_rank_one}
(\mathbf{a}\otimes\mathbf{b}):\mathbb{A}(\mathbf{F}):(\mathbf{a}\otimes\mathbf{b}) \;\geq\; 0
\qquad \forall\,\mathbf{a},\mathbf{b}\in\mathbb{R}^3,
\end{equation}
where $\mathbb{A}_{ijkl}(\mathbf{F}) = \partial^2\Psi / \partial F_{ij}\,\partial F_{kl}$ is the fourth-order elasticity tensor, evaluated at admissible deformations $\mathbf{F}$. As argued in \citet{linden2023neural, neff2015exponentiated, zee1983ordinary}, ellipticity is the practically relevant implication of polyconvexity, and we test it directly because polyconvexity itself is difficult to validate numerically. Note that for incompressible materials the sharp Legendre--Hadamard condition restricts the rank-one perturbations to the isochoric subspace $\mathbf{a}\cdot\mathbf{F}^{-T}\mathbf{b}=0$ \citep{zee1983ordinary}; we deliberately test the stronger, unconstrained condition on the isochoric potential, consistent with polyconvexity as the construction principle targeted by the Creator.

\paragraph{Sampling of $\mathbf{F}$ (isotropic).}
For isotropic, objective potentials, $\Psi(\mathbf{Q}_1\mathbf{F}\mathbf{Q}_2^T) = \Psi(\mathbf{F})$ for all $\mathbf{Q}_1\in\mathcal{SO}(3),\mathbf{Q}_2\in\mathcal{O}(3)$. The rank-one condition at a general $\mathbf{F}$ with directions $(\mathbf{a},\mathbf{b})$ is therefore equivalent to the same condition at $\mathbf{F}=\mathrm{diag}(\lambda_1,\lambda_2,\lambda_3)$ with appropriately rotated directions $(\mathbf{Q}_1\mathbf{a},\mathbf{Q}_2\mathbf{b})$, which still range over all of $\mathbb{R}^3$. Incompressibility further restricts the principal stretches to the surface $\lambda_1\lambda_2\lambda_3=1$, and isotropy makes the condition invariant under permutations of $(\lambda_1,\lambda_2,\lambda_3)$. We therefore sample on the ordered sector $\lambda_1\geq\lambda_2\geq\lambda_3$ with $\lambda_3=(\lambda_1\lambda_2)^{-1}$. The principal stretches are placed on a regular grid in logarithmic stretch space,
\begin{equation}
\log\lambda_1,\log\lambda_2 \in [-L,L],\qquad L=2,
\end{equation}
with $50\times 50$ grid points, so that extension and compression are treated symmetrically. Because the induced stretch $\lambda_3=(\lambda_1\lambda_2)^{-1}$ can reach $e^{2L}$ on this grid, the states are additionally restricted to a problem-dependent range $\max_i|\log\lambda_i|\leq \ell_\mathrm{max}$, with $\ell_\mathrm{max}=2.0$ for the rubber datasets and $\ell_\mathrm{max}=0.8$ for brain tissue. The retained stretches thus extend to $e^{2}\approx 7.4$ for rubber and to $e^{0.8}\approx 2.2$ for brain tissue, in both cases well beyond the range of the respective experimental data, while excluding the extreme states at which the rank-one form becomes dominated by floating-point cancellation. After ordering, deduplication, and this restriction, this yields the validation set $\mathcal{F}_\mathrm{ell}$.

\paragraph{Sampling of $\mathbf{F}$ (transversely isotropic).}
When a fiber is present, the reduction to diagonal form is no longer available: objectivity still removes the left rotation ($\mathbf{F}=\mathbf{R}\mathbf{U}$ implies $\Psi(\mathbf{F})=\Psi(\mathbf{U})$), but the fiber prevents diagonalizing the stretch tensor $\mathbf{U}$ by a material rotation, and the ordering of the in-plane stretches is no longer a symmetry. We therefore sample symmetric gradients whose principal axes are swept relative to the fixed fiber,
\begin{equation}
\mathbf{F} = \mathbf{Q}\,\mathrm{diag}(\lambda_1,\lambda_2,\lambda_3)\,\mathbf{Q}^T,
\qquad \lambda_3=(\lambda_1\lambda_2)^{-1},
\end{equation}
where the orientation $\mathbf{Q}$ ranges over the in-plane rotations $\mathbf{Q}_\mathrm{2D}(\beta)$ at $n_\beta=8$ equispaced angles $\beta\in[0,\pi)$ and, in addition, over the nine three-dimensional rotations of $\mathcal{Q}^+$. The out-of-plane orientations are essential: in-plane rotations keep the $z$-axis a principal direction, so the resulting states never expose the fiber to out-of-plane shear, to which the pseudo-invariant $I_5=\mathrm{cof}(\mathbf{C}):\mathbf{N}$ couples. The stretch pair $(\lambda_1,\lambda_2)$ is placed on a $16\times 16$ logarithmic grid over $[-L,L]$ with $L=1.5$, and the states are then restricted to $\max_i|\log\lambda_i|\leq \ell_\mathrm{max}=0.8$ as for brain tissue, so that all principal stretches -- including the induced out-of-plane stretch $\lambda_3=(\lambda_1\lambda_2)^{-1}$ -- lie in $[e^{-0.8},e^{0.8}]\approx[0.45,2.2]$. This range still exceeds that of the skin data, while avoiding the extreme states at which the rank-one form becomes dominated by floating-point cancellation. The pair is ordered $\lambda_1\geq\lambda_2$; swapping the pair at orientation $\beta$ is equivalent to keeping it at orientation $\beta+\pi/2$, which lies on the grid for the chosen spacing $\pi/8$, so the in-plane sweep covers the swap exactly. For the nine three-dimensional orientations no such companion orientation exists in the set, and the ordering there simply selects one representative of the two in-plane stretch assignments, which we consider sufficient given the frame diversity these rotations add. The out-of-plane stretch $\lambda_3$ is kept separate rather than folded into that ordering. This yields the validation set $\mathcal{F}_\mathrm{ell}^\mathrm{TI}$.

\paragraph{Sampling of $\mathbf{a}$ and $\mathbf{b}$.}
Equation~\eqref{eq_app_rank_one} is homogeneous of degree two in both $\mathbf{a}$ and $\mathbf{b}$, so it is sufficient to sample unit vectors. Furthermore, the substitution $\mathbf{a}\mapsto-\mathbf{a}$ leaves the form invariant, so each direction can be restricted independently to a hemisphere. We use a Fibonacci spiral with $n_\mathrm{dirs}=200$ points covering the upper hemisphere near-uniformly; the same point set serves as both $\mathcal{A}$ and $\mathcal{B}$, and all pairs $(\mathbf{a},\mathbf{b})\in\mathcal{A}\times\mathcal{B}$ are built, including the symmetric pairs $\mathbf{a}=\mathbf{b}$. This direction sampling is identical for the isotropic and transversely isotropic cases. As emphasized in \citet{neff2015exponentiated, zee1983ordinary}, axis-aligned direction pairs probe only the necessary Baker--Ericksen and tension--extension inequalities; the genuinely three-dimensional rank-one condition is exposed only by oblique pairs, which the Fibonacci sampling provides.

\paragraph{Evaluation.}
For each $\mathbf{F}$ in the applicable validation set ($\mathcal{F}_\mathrm{ell}$ or $\mathcal{F}_\mathrm{ell}^\mathrm{TI}$), the elasticity tensor 
$\mathbb{A}(\mathbf{F})$ is computed by nested automatic differentiation 
of $\Psi$ with respect to $\mathbf{F}$ and reshaped into a $9\times 9$ 
matrix $\mathbb{A}_\mathrm{flat}$. For each direction pair, the scalar
\begin{equation}
\mathcal{R}(\mathbf{F},\mathbf{a},\mathbf{b}) \;=\; \mathbf{h}^T\mathbb{A}_\mathrm{flat}(\mathbf{F})\,\mathbf{h},
\qquad \mathbf{h}=\mathrm{vec}(\mathbf{a}\otimes\mathbf{b}),
\end{equation}
is evaluated. The deformation set is processed in mini-batches of $64$
for memory efficiency. Because the entries of $\mathbb{A}$ span many orders of magnitude at strongly deformed states, the rank-one form loses precision to cancellation in single-precision arithmetic, and a sharp sign test would reject provably polyconvex models on round-off noise of relative size $\sim 10^{-6}$. Negativity is therefore judged relative to the per-state magnitude of the form: with $s(\mathbf{F})=\max_{(\mathbf{a},\mathbf{b})}|\mathcal{R}(\mathbf{F},\mathbf{a},\mathbf{b})|$, the validator returns ``failed'' if $\mathcal{R} < -\bigl(\tau_\mathrm{rel}\,s(\mathbf{F}) + \tau_\mathrm{abs}\bigr)$ for any sampled pair, with $\tau_\mathrm{rel}=10^{-3}$ and $\tau_\mathrm{abs}=10^{-4}$ as in the other validators, or if the form evaluates to a non-finite value at any sampled state, and ``passed'' otherwise. Genuine ellipticity violations produce negative values of order one relative to $s(\mathbf{F})$ and are unaffected by this tolerance.

\subsection{Growth condition}\label{app_growth}

Under the incompressibility constraint, every admissible deformation satisfies $J=1$, and the growth condition $\Psi\rightarrow\infty$ as $J\rightarrow 0^+$ or $J\rightarrow\infty$ is therefore trivially fulfilled on the admissible set. The validator returns ``passed'' without any model evaluation.

\subsection{Energy normalization}\label{app_energy_norm}

The model is queried at $\mathbf{F}=\mathbf{I}$, and the validator checks
\begin{equation}
\bigl|\Psi(\mathbf{I})\bigr| \leq \tau.
\end{equation}

\subsection{Stress normalization}\label{app_stress_norm}

Analogously, the model is queried at $\mathbf{F}=\mathbf{I}$, and the validator checks
\begin{equation}
\bigl\|\mathbf{P}(\mathbf{I})\bigr\|_\infty \leq \tau.
\end{equation}
The check itself is identical for isotropic and transversely isotropic materials; only the means by which a stress-free reference is achieved during model construction differs, as discussed in Section~\ref{sec_Background_constraints}.

\subsection{Non-negativity of the strain energy}\label{app_nonneg}

Finally, the validator evaluates $\Psi$ at every sampled deformation gradient and checks
\begin{equation}
\Psi(\mathbf{F}) \;\geq\; -\tau.
\end{equation}
For isotropic materials, the sample is $\mathcal{F}_\mathrm{base}$. For transversely isotropic materials, where non-negativity cannot be guaranteed by construction (cf.~Section~\ref{sec_Background_constraints}) and this check is therefore the decisive one, the sample is $\mathcal{F}_\mathrm{base}^\mathrm{TI}$ additionally extended by the rotated symmetric deformation gradients of the ellipticity construction, used here without the stretch restriction, i.e., on the full grid with $L=1.5$. These states reach substantially stronger fiber compression (squared fiber stretches down to $I_4\approx e^{-3}$) than both the base set and the restricted ellipticity set $\mathcal{F}_\mathrm{ell}^\mathrm{TI}$. A small negative tolerance is admitted to absorb floating-point noise around $\Psi=0$ in the reference configuration.

\newpage

\section{Adherence of generated models to physical constraints in absolute numbers} \label{sec_Appendix_verifications_absolute}

\begin{figure}[h]
  \centering
  \includegraphics[width=\linewidth]{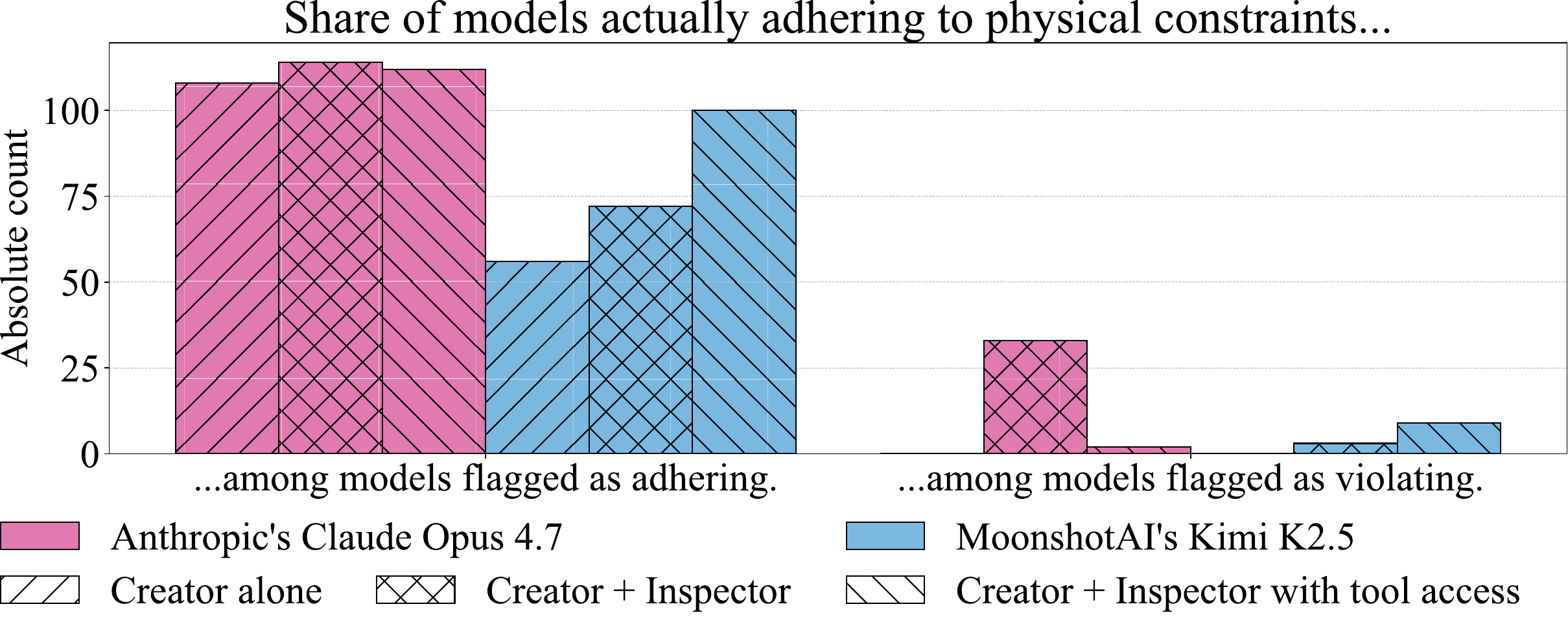}
  \caption{Adherence of generated models to physical constraints in absolute counts. Same layout as Figure~\ref{fig_predictions_cann_verifications_pct}, but reporting absolute numbers of models rather than relative frequencies. The lower counts on the flagged-as-violating side reflect that only a random subset of violating models is exported for analysis, whereas every refinement round exports exactly one inspector-approved model.}
  \label{fig_predictions_cann_verifications_abs}
\end{figure}

Figure~\ref{fig_predictions_cann_verifications_pct} in Section~\ref{sec_Results_agents} reports the adherence of the generated models to the physical constraints in relative terms, which is the most informative view for judging how much a user can trust a model returned by the pipeline. For completeness, we also provide the underlying absolute counts in Figure~\ref{fig_predictions_cann_verifications_abs}. Two points are worth noting when interpreting these numbers. First, every refinement round exports exactly one inspector-approved model, so the totals on the flagged-as-adhering side directly reflect the number of runs. Second, models flagged by the inspector as violating are not exported by default; we export only a random subset of them for the purpose of this analysis, which is why the totals on the flagged-as-violating side are substantially lower. This imbalance is a consequence of our experimental design and does not affect the relative frequencies discussed in Section~\ref{sec_Results_agents}.

\newpage

\section{Evaluation of models generated with minimal prompts} \label{sec_Appendix_minimal_prompts}

\begin{figure}[h]
  \centering
  \includegraphics[width=\linewidth]{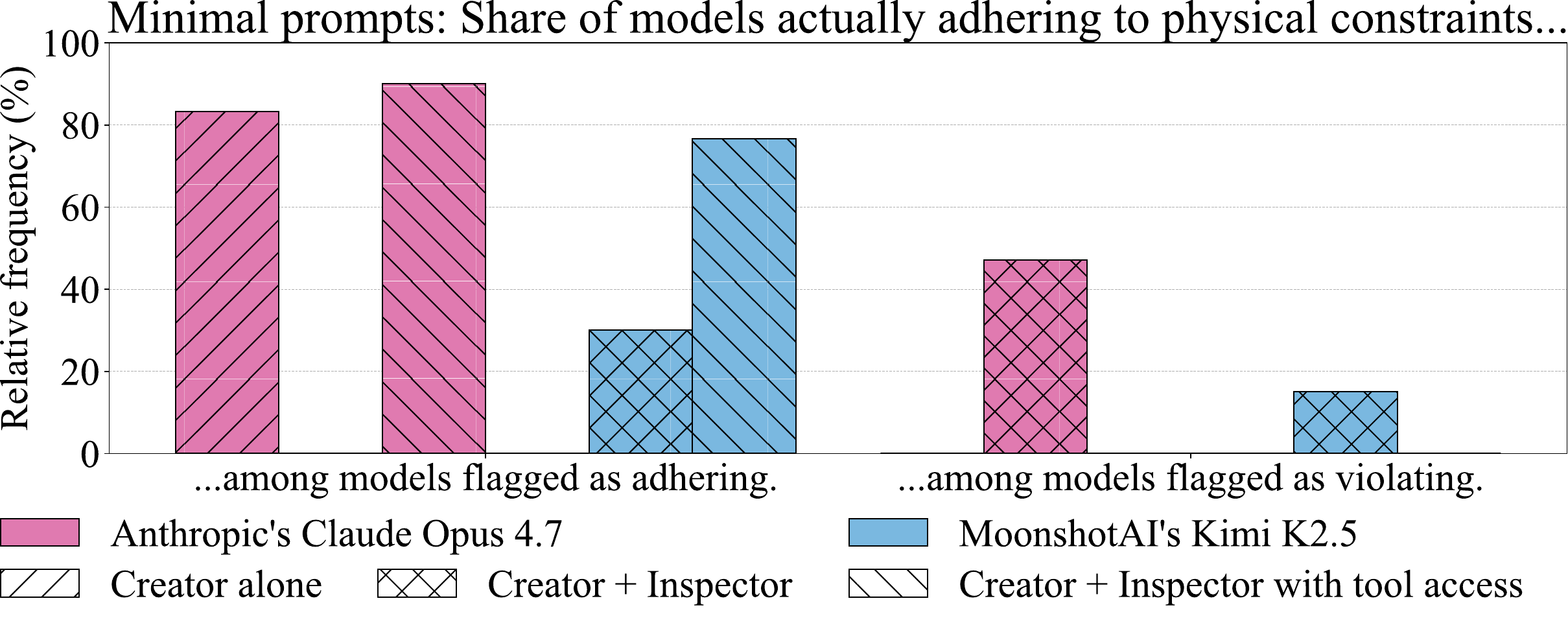}
  \caption{Adherence of generated models to physical constraints when LLMs are prompted minimally.}
  \label{fig_predictions_cann_verifications_minimal}
\end{figure}

\begin{figure}[h]
  \centering
  \includegraphics[width=\linewidth]{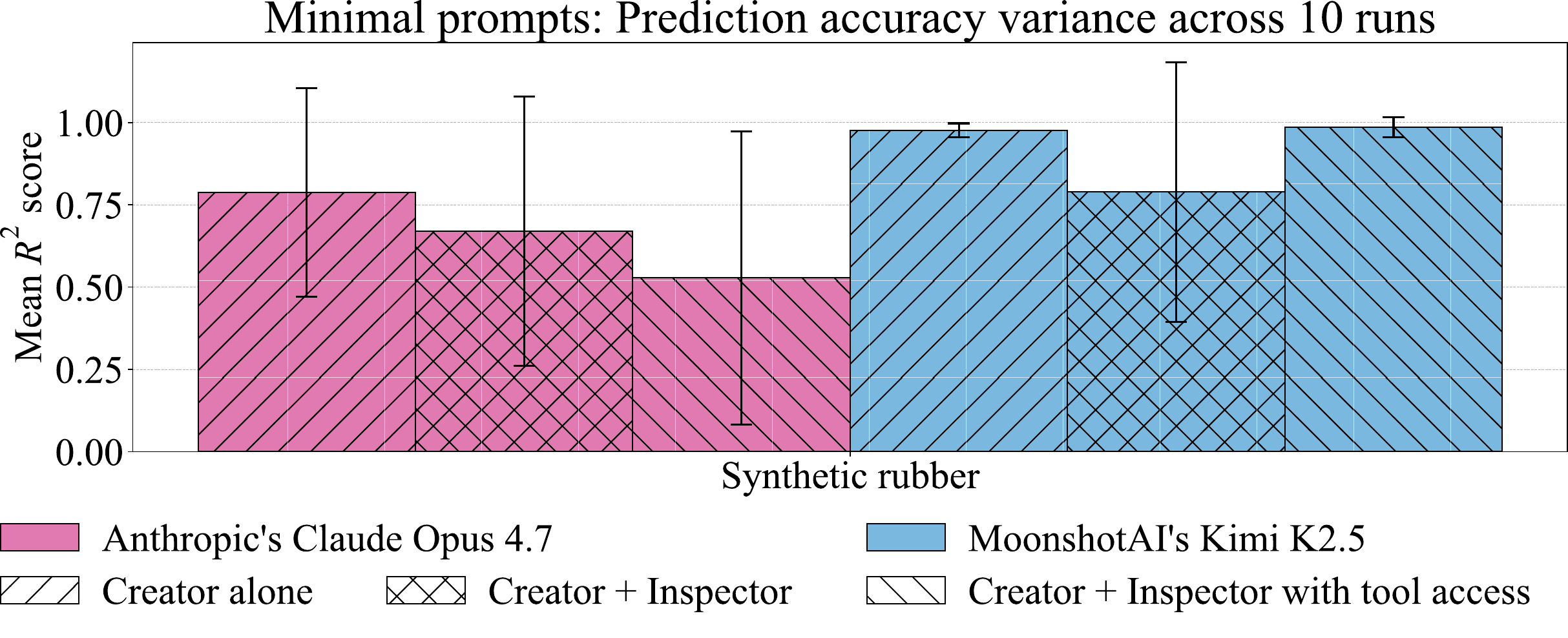}
  \caption{Reliability of the achieved accuracy across runs when LLMs are prompted minimally. Mean $R^2$ score of model performance, averaged over the three loading conditions of the synthetic rubber dataset and over ten independent runs and showing also standard deviation. Negative average R2 scores were clipped to zero to keep extreme outliers from dominating the plot.}
  \label{fig_predictions_accuracy_variance_minimal}
\end{figure}

In our main experiments, both agents are informed in detail about the physical constraints: the Creator is told what each constraint requires, and the Inspector additionally receives practical hints on how adherence can be checked. To measure how much of the system's reliability rests on this information, we repeated the experiments on the synthetic rubber dataset with minimal prompts, which merely name the nine constraints without specifying what exactly they demand or how they can be enforced. Figure~\ref{fig_predictions_cann_verifications_minimal} shows the consequence: under-informed agents are consistently less reliable. Every configuration falls below its fully informed counterpart on the same dataset — the Opus-powered configurations drop from 90\% to 83\% (Creator alone), from 100\% to 0\% (Creator with Inspector), and from 97\% to 90\% (with tool access), and the Kimi-powered ones from 47\% to 0\%, from 57\% to 30\%, and from 87\% to 77\%. The accuracies in Figure~\ref{fig_predictions_accuracy_variance_minimal} become noticeably more erratic as well, particularly for the Opus-powered configurations. How precisely the agents are informed about the constraints thus directly limits what the pipeline can deliver: the paradigm provides the division of labor, but the prompts provide the domain knowledge.

\begin{figure}[h!]
  \centering
  \includegraphics[width=\linewidth]{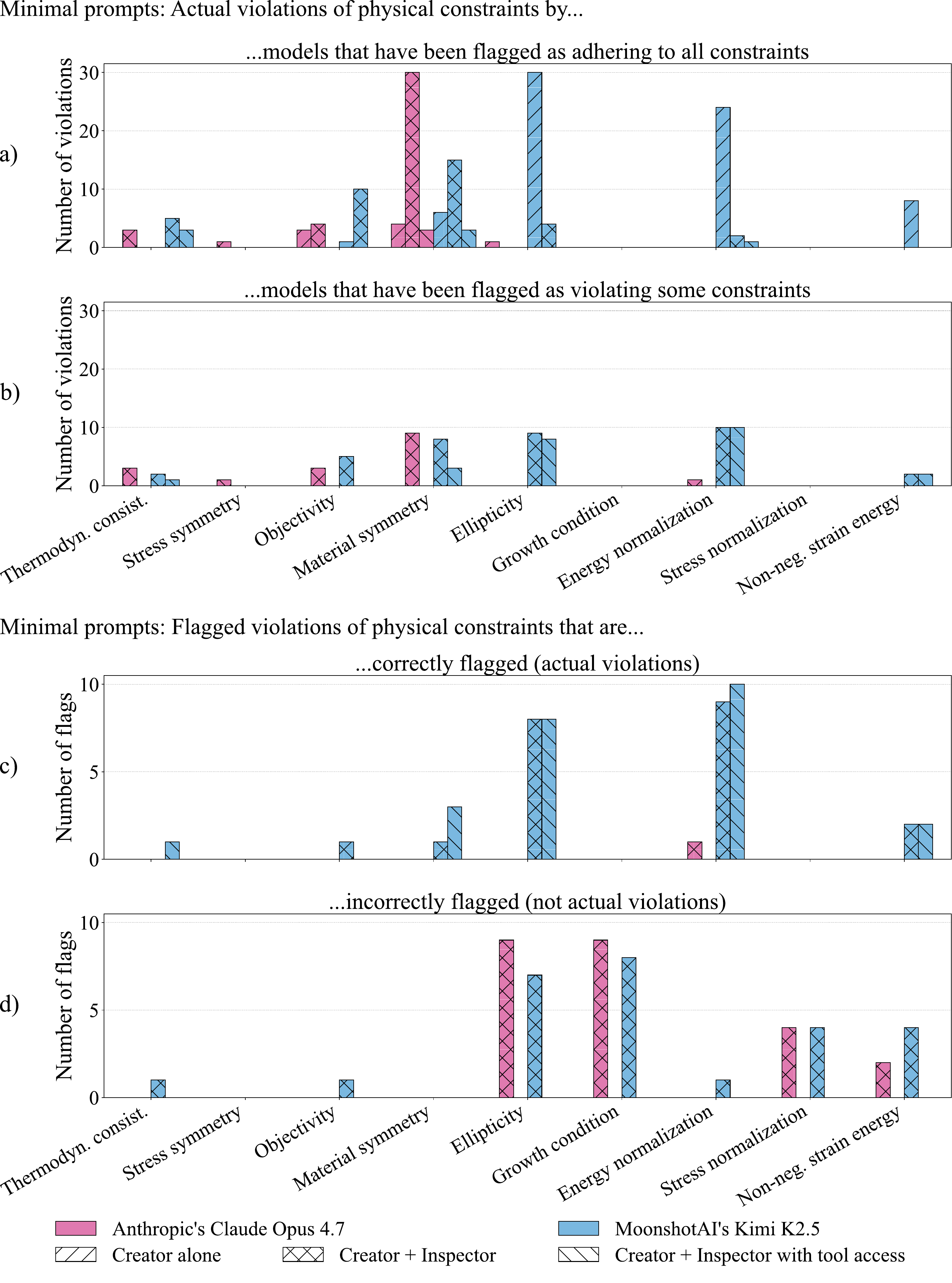}
  \caption{Distribution of physical constraint violations by type, separated into actual vs. flagged and correct vs. incorrect when LLMs are prompted minimally.}
  \label{fig_predictions_constraint_types_minimal}
\end{figure}

The most striking result of this ablation is the complete failure of the Opus-powered Creator–Inspector pair: not a single one of its 30 exported models withstands the external validation, even though the Opus Creator alone reaches 83\%. Figure~\ref{fig_predictions_constraint_types_minimal} reveals the mechanism behind this collapse. In the well-informed setting of Section~\ref{sec_Results_verification}, we found the Opus Inspector to be a well-adjusted skeptic that would rather doubt a fulfilled constraint than accept a violated one — a favorable trait there. Under minimal prompts, the same skepticism turns into a trap. Chart d) shows that the Inspector doubts constraints the unmodified models actually satisfy, most often ellipticity, the growth condition, stress normalization, and non-negativity of the strain energy. The Creator then answers these doubts in essentially every round with the same remedy, a volumetric term based on $\det(\mathbf{F})$, and this remedy breaks material symmetry: as Chart a) shows, all 30 models that the pair exports as adhering violate exactly this constraint. The trap is specific to this backbone under this set of prompts. For Kimi, the Inspector's interventions fix far more than they break, lifting adherence from 0\% to 30\%, and a systematic collapse like the one observed for Opus appears nowhere else. We therefore read this failure as a model-specific behavior under under-informed prompting, not as a weakness of the Creator–Inspector paradigm itself. Consistently, the failure disappears as soon as the Inspector's beliefs are replaced by measurements: with tool access, the doubts are resolved numerically, and the Opus-powered pair reaches 90\%.

\newpage

\newpage

This ablation also holds a broader lesson. The core motivation of our pipeline is accessibility and flexibility: constitutive models are generated on demand and tailored specifically to the user's problem, so no individual model persists as the defining artifact of the approach. The lasting artifact is instead the orchestration and prompting of the agents that produce these models. The experiments in this appendix show how much rests on precisely this part: orchestration and prompting should not be underestimated — they form a central and integral core of the pipeline.

\newpage

\section{Model implementations generated for the experimental rubber dataset} \label{sec_Appendix_exemplary_canns}

To give a concrete impression of the artifacts produced by our pipeline, we show below the Python implementations of the two CANNs whose predictions on the experimental rubber dataset are shown in Figures~\ref{fig_predictions_curves_opus} and \ref{fig_predictions_curves_kimi}. Both were generated by the Creator-Inspector pipeline without tool access and correspond, in each case, to the most accurate adherent model out of ten independent runs — the Opus-powered pair in Listing~\ref{lst_opus_cann} and the Kimi-powered pair in Listing~\ref{lst_kimi_cann}.

\vspace{10pt}

\captionsetup{type=listing2} % tell 'caption' the caption type
\captionof{listing2}{Model implementation for the experimental rubber dataset generated by the Creator--Inspector pipeline powered by Claude Opus 4.7 (best-out-of-ten adherent run, no tool access).}
\label{lst_opus_cann}
\lstset{language=Python, escapeinside={(*@}{@*)}}
\begin{lstlisting}


import tensorflow as tf


class NonNegative(tf.keras.constraints.Constraint):
    def __call__(self, w):
        return tf.nn.relu(w)


class CANNBlock(tf.keras.layers.Layer):
    """
    Polyconvex building block.

    Input: x = (invariant - 3) >= 0, shape (batch, 1).
    Output: non-negative scalar psi contribution, shape (batch, 1).

    All activations are:
      * convex,
      * monotonically increasing,
      * C^2 (or smoother),
      * zero at x = 0 (so psi(C=I)=0; explicit subtraction also applied).

    Neurons per activation type, each with its own non-negative inner and
    outer weights. The inner weights act inside the nonlinearity; the outer
    weights scale the result.
    """

    def __init__(self, num_neurons=2, **kwargs):
        super().__init__(**kwargs)
        self.num_neurons = num_neurons

    def build(self, input_shape):
        n = self.num_neurons

        # Modest initializers to avoid blow-ups; spread scales over neurons.
        init_in_s = tf.keras.initializers.RandomUniform(0.05, 0.3)
        init_in_m = tf.keras.initializers.RandomUniform(0.1, 0.5)
        init_out  = tf.keras.initializers.RandomUniform(0.02, 0.15)
        init_out_small = tf.keras.initializers.RandomUniform(1e-3, 2e-2)

        # --- Linear: f(y) = y (Neo-Hookean-like) ---
        self.w_in_lin = self.add_weight(
            name="w_in_lin", shape=(n,), initializer=init_in_s,
            constraint=NonNegative(), trainable=True)
        self.w_out_lin = self.add_weight(
            name="w_out_lin", shape=(n,), initializer=init_out,
            constraint=NonNegative(), trainable=True)

        # --- Quadratic: f(y) = y^2 ---
        self.w_in_sq = self.add_weight(
            name="w_in_sq", shape=(n,), initializer=init_in_s,
            constraint=NonNegative(), trainable=True)
        self.w_out_sq = self.add_weight(
            name="w_out_sq", shape=(n,), initializer=init_out_small,
            constraint=NonNegative(), trainable=True)

        # --- Softplus-based: f(y) = softplus(y) - log(2) ---
        # f(0)=0, f'(0)=1/2>0, f''>=0, smooth.
        self.w_in_sp = self.add_weight(
            name="w_in_sp", shape=(n,), initializer=init_in_m,
            constraint=NonNegative(), trainable=True)
        self.w_out_sp = self.add_weight(
            name="w_out_sp", shape=(n,), initializer=init_out,
            constraint=NonNegative(), trainable=True)

        # --- Squared softplus: f(y) = (softplus(y) - log(2))^2 ---
        # Convex, monotonically increasing for y>=0, smooth, f(0)=0.
        self.w_in_sp2 = self.add_weight(
            name="w_in_sp2", shape=(n,), initializer=init_in_m,
            constraint=NonNegative(), trainable=True)
        self.w_out_sp2 = self.add_weight(
            name="w_out_sp2", shape=(n,), initializer=init_out_small,
            constraint=NonNegative(), trainable=True)

        # --- Exponential-like (Demiray/Holzapfel): f(y) = exp(y) - 1 - y ---
        # f(0)=0, f'(y)=exp(y)-1>=0 for y>=0, f''(y)=exp(y)>0, smooth.
        self.w_in_exp = self.add_weight(
            name="w_in_exp", shape=(n,),
            initializer=tf.keras.initializers.RandomUniform(0.05, 0.25),
            constraint=NonNegative(), trainable=True)
        self.w_out_exp = self.add_weight(
            name="w_out_exp", shape=(n,),
            initializer=tf.keras.initializers.RandomUniform(1e-4, 5e-3),
            constraint=NonNegative(), trainable=True)

        super().build(input_shape)

    def call(self, x):
        # x shape: (batch, 1). Weight shape: (n,). Broadcast -> (batch, n).

        # Linear
        y_lin = self.w_in_lin * x
        a_lin = y_lin
        out_lin = tf.reduce_sum(self.w_out_lin * a_lin, axis=-1, keepdims=True)

        # Quadratic
        y_sq = self.w_in_sq * x
        a_sq = tf.square(y_sq)
        out_sq = tf.reduce_sum(self.w_out_sq * a_sq, axis=-1, keepdims=True)

        # Softplus shifted
        y_sp = self.w_in_sp * x
        # Limit argument to avoid overflow (softplus is linear for large y).
        y_sp = tf.clip_by_value(y_sp, 0.0, 30.0)
        a_sp = tf.nn.softplus(y_sp) - tf.math.log(2.0)
        out_sp = tf.reduce_sum(self.w_out_sp * a_sp, axis=-1, keepdims=True)

        # Squared softplus
        y_sp2 = self.w_in_sp2 * x
        y_sp2 = tf.clip_by_value(y_sp2, 0.0, 30.0)
        a_sp2 = tf.square(tf.nn.softplus(y_sp2) - tf.math.log(2.0))
        out_sp2 = tf.reduce_sum(self.w_out_sp2 * a_sp2, axis=-1, keepdims=True)

        # Exponential-like
        y_exp = self.w_in_exp * x
        # Strict clipping to keep things numerically safe.
        y_exp = tf.clip_by_value(y_exp, 0.0, 10.0)
        a_exp = tf.exp(y_exp) - 1.0 - y_exp
        out_exp = tf.reduce_sum(self.w_out_exp * a_exp, axis=-1, keepdims=True)

        return out_lin + out_sq + out_sp + out_sp2 + out_exp


class CANN(tf.keras.Model):
    def __init__(self, num_neurons=2, **kwargs):
        super().__init__(**kwargs)
        self.block_I1 = CANNBlock(num_neurons=num_neurons, name="block_I1")
        self.block_I2 = CANNBlock(num_neurons=num_neurons, name="block_I2")

    def psi_from_F(self, F):
        # Right Cauchy-Green tensor
        C = tf.matmul(F, F, transpose_a=True)
        I1 = tf.linalg.trace(C)
        C2 = tf.matmul(C, C)
        I2 = 0.5 * (tf.square(I1) - tf.linalg.trace(C2))

        I1 = tf.reshape(I1, (-1, 1))
        I2 = tf.reshape(I2, (-1, 1))

        # For incompressible isotropic materials I1, I2 >= 3. Clamp noise.
        x1 = tf.nn.relu(I1 - 3.0)
        x2 = tf.nn.relu(I2 - 3.0)

        psi_raw = self.block_I1(x1) + self.block_I2(x2)

        # Explicit energy normalization: psi(C=I) = 0.
        zero_in = tf.zeros_like(x1)
        psi0 = self.block_I1(zero_in) + self.block_I2(zero_in)
        psi = psi_raw - psi0
        return psi

    def call(self, F):
        with tf.GradientTape() as tape:
            tape.watch(F)
            psi = self.psi_from_F(F)
        dpsi_dF = tape.gradient(psi, F)

        # P = dpsi/dF - p * F^{-T}, enforcing P[2,2] = 0 (plane-stress-like in
        # the 3-direction). This determines the Lagrange multiplier p.
        F_invT = tf.linalg.inv(tf.linalg.matrix_transpose(F))
        num = dpsi_dF[:, 2, 2]
        den = F_invT[:, 2, 2]
        den_safe = tf.where(tf.abs(den) < 1e-8,
                            tf.ones_like(den) * 1e-8, den)
        p = num / den_safe
        p = tf.reshape(p, (-1, 1, 1))

        P = dpsi_dF - p * F_invT
        return {"P": P, "Psi": psi}


def build_cann_model():
    model = CANN(num_neurons=2)
    # Force a build with a dummy input so all weights are created.
    dummy_F = tf.eye(3, batch_shape=[1], dtype=tf.float32)
    _ = model(dummy_F)
    return model


\end{lstlisting}

\newpage

\captionsetup{type=listing2} % tell 'caption' the caption type
\captionof{listing2}{Model implementation for the experimental rubber dataset generated by the Creator--Inspector pipeline powered by Kimi K2.5 (best-out-of-ten adherent run, no tool access).}
\label{lst_kimi_cann}
\lstset{language=Python, escapeinside={(*@}{@*)}}
\begin{lstlisting}


import tensorflow as tf


class CANN(tf.keras.Model):
    def __init__(self, **kwargs):
        """ 
        Args:
            kwargs: Additional keyword arguments for the model initialization.
        """
        super(CANN, self).__init__(**kwargs)
        
        # Define a convex neural network architecture
        # Using non-negative weights and convex, monotonically increasing activations (Softplus)
        hidden_units = [10, 10]  # Default architecture, can be adjusted
        
        self.hidden_layers = []
        for units in hidden_units:
            self.hidden_layers.append(
                tf.keras.layers.Dense(
                    units,
                    activation='softplus',
                    kernel_constraint=tf.keras.constraints.NonNeg(),
                    dtype='float32'
                )
            )
        
        # Output layer: scalar strain energy density
        self.psi_output = tf.keras.layers.Dense(
            1,
            activation='softplus',
            kernel_constraint=tf.keras.constraints.NonNeg(),
            dtype='float32'
        )

    def _compute_invariants(self, F):
        """Compute the first and second invariants of the right Cauchy-Green tensor C."""
        # C = F^T * F
        C = tf.matmul(F, F, transpose_a=True)
        
        # First invariant: I1 = tr(C)
        I1 = tf.linalg.trace(C)
        
        # Second invariant: I2 = 0.5 * (tr(C)^2 - tr(C^2))
        trace_C2 = tf.linalg.trace(tf.matmul(C, C))
        I2 = 0.5 * (tf.square(I1) - trace_C2)
        
        # Reshape to (batch_size, 1)
        I1 = tf.expand_dims(I1, axis=-1)
        I2 = tf.expand_dims(I2, axis=-1)
        
        return I1, I2

    def psi_from_F(self, F):
        """
        Pure forward computation of psi from F. No tf.GradientTape inside.

        Args:
            F (tf.Tensor): Input tensor containing the deformation gradient F. Shape: (batch_size,3,3).
        Returns:
            tf.Tensor: Strain energy density psi predicted from the deformation gradient F. Shape: (batch_size,1)
        """
        # Compute invariants I1 and I2
        I1, I2 = self._compute_invariants(F)
        
        # Prepare input features (batch_size, 2)
        features = tf.concat([I1, I2], axis=-1)
        
        # Forward pass through the network
        x = features
        for layer in self.hidden_layers:
            x = layer(x)
        psi_raw = self.psi_output(x)
        
        # Energy normalization: subtract psi evaluated at C = I (where I1=3, I2=3)
        I1_id = tf.constant([[3.0]], dtype=tf.float32)
        I2_id = tf.constant([[3.0]], dtype=tf.float32)
        features_id = tf.concat([I1_id, I2_id], axis=-1)
        
        x_id = features_id
        for layer in self.hidden_layers:
            x_id = layer(x_id)
        psi_id = self.psi_output(x_id)
        
        psi = psi_raw - psi_id
        
        return psi

    def call(self, F):
        """
        Uses method psi_from_F and tf.GradientTape to predict both psi and P from F.

        Args:
            F (tf.Tensor): Input tensor containing the deformation gradient F. Shape: (batch_size,3,3)
        Returns:
            dict: Dictionary with keys "P" and "Psi":
                  - "P": First Piola-Kirchhoff stress tensor. Shape: (batch_size,3,3).
                  - "Psi": Strain energy density. Shape: (batch_size,1).
        """
        with tf.GradientTape() as tape:
            tape.watch(F)
            psi = self.psi_from_F(F)
        
        # Compute isochoric part of P = psi/F
        P_iso = tape.gradient(psi, F)
        
        # Handle potential None gradient (should not happen with valid operations)
        if P_iso is None:
            P_iso = tf.zeros_like(F)
        
        # Compute volumetric part based on hydrostatic pressure p
        # For incompressible materials: P = P_iso - p * F^{-T}
        F_inv = tf.linalg.inv(F)
        F_inv_T = tf.transpose(F_inv, perm=[0, 2, 1])
        
        # Determine p from boundary condition P_{33} = 0 (index 2,2)
        # P_{33} = P_iso_{33} - p * (F^{-T})_{33} = 0
        # => p = P_iso_{33} / (F^{-T})_{33}
        P_iso_33 = P_iso[:, 2, 2]
        F_inv_T_33 = F_inv_T[:, 2, 2]
        
        # Add small epsilon to avoid division by zero
        eps = 1e-8
        p = P_iso_33 / (F_inv_T_33 + eps)
        
        # Reshape p for broadcasting: (batch_size, 1, 1)
        p = tf.reshape(p, [-1, 1, 1])
        
        # Total first Piola-Kirchhoff stress
        P = P_iso - p * F_inv_T
        
        return {"P": P, "Psi": psi}


def build_cann_model():
    """ Builds a CANN model for predicting the strain energy density psi and the first Piola Kirchhoff stress P from the deformation gradient F for hyperelastic, incompressible, isotropic materials.
    Args:
        -- None --
    Returns:
        CANN: A CANN model instance (subclass of tf.keras.Model).
    """
    return CANN()


\end{lstlisting}

\end{document}